\documentclass[journal]{IEEEtran}
\usepackage{array}
\usepackage[caption=false,font=normalsize,labelfont=sf,textfont=sf]{subfig}
\usepackage{textcomp}
\usepackage{stfloats}
\usepackage{url}
\usepackage{verbatim}
\usepackage{booktabs}
\usepackage{graphicx}
\usepackage[caption=false]{subfig}
\usepackage{multirow}
\usepackage{tabularx}
\usepackage{url}
\usepackage{amsmath}
\usepackage{amssymb}
\usepackage{tikz}
\usepackage{pgf, pgfsys, pgffor}
\usepackage{float}
\usepackage[caption=false]{subfig}
\usepackage[switch]{lineno}
\usepackage[linesnumbered,ruled]{algorithm2e}
\newcolumntype{Y}{>{\centering\arraybackslash}X}
\usepackage[pagebackref=true,breaklinks=true,letterpaper=true,colorlinks,citecolor=blue,linkcolor=blue,bookmarks=false]{hyperref}

\ifCLASSINFOpdf
\else
\fi
\graphicspath{{./figure}}
\begin{document}
%
\title{End-to-End Human Instance Matting}
%
%
%
\author{Qinglin Liu,
Shengping Zhang,
        Quanling Meng,
        Bineng Zhong,
        Peiqiang Liu,
        and~Hongxun Yao
\thanks{\emph{Corresponding author: Shengping Zhang.} }
\thanks{Qinglin Liu, Shengping Zhang, Quanling Meng, and Hongxun Yao are with Harbin Institute of Technology, Harbin 150001, China (e-mail: qinglin.liu@outlook.com; s.zhang@hit.edu.cn; quanling.meng@hit.edu.cn; 	h.yao@hit.edu.cn).}
\thanks{ Bineng Zhong is with Guangxi Normal University, Guilin 541004, China (e-mail:bnzhong@hqu.edu.cn).}
\thanks{ Peiqiang Liu  is with Shandong Technology and Business University, Yantai 264003, China (e-mail:liupq@126.com).}
}

%
%

\markboth{Journal of \LaTeX\ Class Files,~Vol.~14, No.~8, August~2015}%
{Shell \MakeLowercase{\textit{et al.}}: Bare Demo of IEEEtran.cls for IEEE Journals}
%



\maketitle

\begin{abstract}
Human instance matting aims to estimate an alpha matte for each human instance in an image, which is extremely challenging and has rarely been studied so far.
Despite some efforts to use instance segmentation to generate a trimap for each instance and apply trimap-based matting methods, the resulting alpha mattes are often inaccurate due to inaccurate segmentation.
In addition, this approach is computationally inefficient due to multiple executions of the matting method.
To address these problems, this paper proposes a novel End-to-End Human Instance Matting (E2E-HIM) framework for simultaneous multiple instance matting in a more efficient manner.
Specifically, a general perception network first extracts image features and decodes instance contexts into latent codes.
Then, a united guidance network exploits spatial attention and semantics embedding to generate united semantics guidance, which encodes the locations and semantic correspondences of all instances.
Finally, an instance matting network decodes the image features and united semantics guidance to predict all instance-level alpha mattes.
In addition, we construct a large-scale human instance matting dataset (HIM-100K) comprising  over 100,000 human images with instance alpha matte labels.
Experiments on HIM-100K demonstrate the proposed E2E-HIM outperforms the existing methods on human instance matting with $50 \%$ lower errors and $5 \times$ faster speed (6 instances in a 640 $\times$ 640 image).
Experiments on the PPM-100, RWP-636, and P3M datasets demonstrate that E2E-HIM also achieves competitive performance on traditional human matting. %
\end{abstract}


\vspace{-1pt}
\begin{figure*}[!t]
	\begin{center}
\includegraphics[width=1.\linewidth]{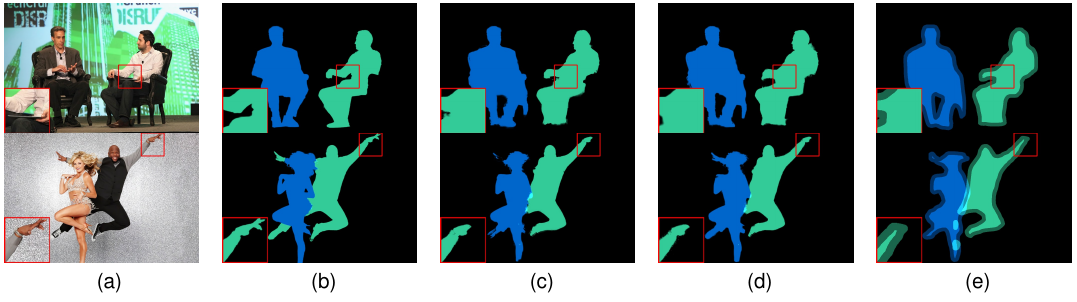}
\end{center}
 \vspace{-5pt}
\caption{Results of instance-level alpha mattes estimated by the proposed E2E-HIM and two state-of-the-art methods.  (a) Input images. (b) Alpha mattes estimated by our E2E-HIM.(c) Alpha mattes estimated by ISSMatting~\cite{instmattcrv} (Mask R-CNN~\cite{he2020mask} + DIM~\cite{xu2017deep}). (d) Alpha mattes estimated by InstMatt~\cite{sun2022instmatt} (Mask R-CNN + FBAMatting~\cite{forte2020fbamatting}).  (e) Trimaps used by ISSMatting and InstMatt, which are generated from the instance segmentation of Mask R-CNN.}
 \vspace{-3pt}
\label{fig:badseg}
\end{figure*}

\section{Introduction}
\IEEEPARstart{H}{uman} matting aims to estimate the alpha matte (opacity) of the humans (usually only one person) in an image, which has many potential applications, such as image editing~\cite{zhu2017fast,chen2018semantic} and video post-production~\cite{Wang_2021_ICCV,zhang2021attention}.
According to whether auxiliary guidance such as a trimap (specifying the foreground, background, and transition regions in an image) is used or not, existing matting methods can be roughly grouped into two categorizations: trimap based matting~\cite{levin2008a,chen2013knn, xu2017deep,lu2019indices,li2020natural,forte2020fbamatting,yu2020mask,Yu_2021_ICCV} and automatic matting~\cite{zhu2017fast,Zhang2019A,Yu_2021_ICCV,ke2020is,li2022matting}.
Due to the extensive labor of annotating trimaps, automatic matting methods have been attracting increasing attention, which automatically generate trimaps based on instance segmentation~\cite{instmattcrv,sun2022instmatt} or directly estimate alpha mattes in an end-to-end manner~\cite{Shen2016Deep,Zhang2019A,ke2020is,2021Privacy,li2022matting}.

The rapid development of live streaming and short video applications has created a growing demand for personalized editing of human instances. 
Various personalized editing tasks, such as body shape modifications~\cite{ren2022structure,chen2022single} and skin toning~\cite{li2015deep,Velusamy_2020_CVPR_Workshops}, require the alpha matte of each human instance in the image.
However, traditional interactive matting methods~\cite{xu2017deep,yu2020mask,wei2021improved,yang2022unified,ding2022deep} rely on extensive  manual intervention, which makes them time-consuming and labor-intensive when dealing with a large number of images and videos.
This paper focuses on automatically solving this problem, which we refer to as \emph{human instance matting}.
Formally, let $\boldsymbol{I}\in\mathbb{R}^{3\times H\times W}$ be an input image with $X$ human instances, where ${H}$ and $W$ are the height and width of the image, respectively.
Human instance matting aims to estimate alpha mattes $\mathcal{A}=\{\boldsymbol{A}^1, \boldsymbol{A}^2, \ldots, \boldsymbol{A}^X\}$, where $\boldsymbol{A}^x\in \mathbb{R}^{ H \times W }$ is the alpha matte for the $x$-{th} instance such that
\begin{equation}\label{eq:indicator}
\boldsymbol{I} =  \sum_{x=1}^X \boldsymbol{A}^x \odot \boldsymbol{F}^x+(1-\sum_{x=1}^X\boldsymbol{A}^x) \odot \boldsymbol{B}
\end{equation}
where $\boldsymbol{F}^x \in\mathbb{R}^{3\times H\times W}$ and $\boldsymbol{B} \in\mathbb{R}^{3\times H\times W}$ are the foreground for the $x$-{th} instance  and background.
$\odot$ is the broadcasted element-wise multiplication.

Compared with traditional human matting that estimates the alpha matte of all humans in an image, human instance matting estimates the alpha matte of each human instance, which is more challenging and has rarely been studied so far.
To perform instance-level matting, several efforts~\cite{instmattcrv,sun2022instmatt} have been devoted to exploiting instance segmentation~\cite{he2020mask} to first generate a trimap for each instance and then individually execute  an existing trimap-based matting method~\cite{xu2017deep,forte2020fbamatting,zhang2020mask} for each human instance.
However, such a two-stage solution has several disadvantages.
\emph{First}, existing instance segmentation methods~\cite{he2020mask, wang2020solo,dong2021solq} usually generate low-resolution segmentation masks (e.g.,  $112 \times 112$ by Transfiner~\cite{transfiner}, $160 \times 160$ by SOLO~\cite{wang2020solo}), which are not accurate enough to distinguish fine-grained details (e.g., curved limbs) and therefore causes the generated trimaps to be very coarse, especially in complex poses. 
\emph{Second}, the generation of the trimaps and estimation of alpha mattes are independently performed, which cannot be jointly optimized and therefore limit the matting performance.
As shown in Figure~\ref{fig:badseg}, once the trimaps generated from segmentation are coarse, the obtained human instance alpha mattes are not accurate.
\emph{Third}, the matting method is performed individually for each human instance, which leads to expensive computational consumption (e.g., the scheme~\cite{sun2022instmatt} that combines Mask R-CNN and FBAMatting~\cite{forte2020fbamatting} to infer a $640 \times 640$ image with 6 instances costs 1,500 GFlops and 0.5 seconds on an Nvidia RTX 2080Ti GPU).

To address these problems, this paper proposes a novel End-to-End Human Instance Matting (E2E-HIM) framework for simultaneously estimating all instance-level alpha mattes in a more efficient way, which contains three jointly optimized sub-networks: a general perception network, a united guidance network, and an instance matting network.
Specifically, a general perception network adopts a hybrid transformer to first extract image features and then decode instance contexts into latent codes.
To help aggregate image features for instance matting, a united guidance network exploits spatial attention and semantics embedding to generate the united semantics guidance, which encodes both the locations and semantic correspondences of all human instances.
To perform instance matting, an instance matting network dynamically decodes the image features and united semantics guidance to simultaneously predict all instance-level alpha mattes.
In addition, we construct a large-scale Human Instance Matting (HIM-100K) dataset, which contains more than 100,000 images with about 326,000 human instances.
Each human instance in an image has three annotations including the alpha matte, segmentation mask, and bounding box.
Experiments on the HIM-100K dataset demonstrate that the proposed E2E-HIM outperforms existing human instance matting methods.
In particular, E2E-HIM using the ResNet-50~\cite{he2016deep} backbone achieves $50 \%$ error reduction and $5 \times$ faster speed while only consuming 305 GFlops to estimate all instance-level alpha mattes in a $640 \times 640$ image.
Experimental results on PPM-100~\cite{ke2020is}, RWP-636~\cite{zhang2020mask}, and P3M~\cite{2021Privacy} show that E2E-HIM also achieves competitive performance on human matting.
The contributions of our work can be summarized as follows:
\begin{itemize}
\item We propose the first End-to-End Human Instance Matting (E2E-HIM) to our best knowledge,  which jointly optimizes instance context extraction and matting in a unified network and therefore significantly improves the performance while achieving faster speed compared with instance segmentation based two-stage matting methods.
\item We introduce a united guidance network to encode both the locations and semantic correspondences of all human instances to obtain a united semantics guidance, which discriminates human instances and  facilitates the prediction of instance-level alpha mattes simultaneously.

\item We collect a large-scale human instance matting dataset, named HIM-100K, containing more than 100,000 human images with about 326,000 human instances. Each instance has three annotations including the alpha matte,  segmentation mask, and  bounding box, which is valuable for future research in human matting.
\item Experimental results on HIM-100K demonstrate the proposed method outperforms the state-of-the-art human instance matting methods. Moreover, experiments  on PPM-100, RWP-636, and P3M also demonstrate that E2E-HIM achieves competitive performance on human matting.
\end{itemize}

\section{Related Work}
In this section, we review the instance segmentation methods and image matting methods that are related to our work.

\noindent\textbf{Instance Segmentation.}
Instance segmentation aims to predict the category and mask of each object in an image.
Existing methods can be roughly categorized into three groups: top-down, bottom-up, and direct methods.

%
Traditional top-down methods use Mask R-CNN~\cite{he2020mask} and PANet~\cite{liu2018path} to predict the candidate regions and then estimate the instance segmentation inside the regions.
Newly emerging methods~\cite{zhang2020mask,bolya2019yolact,9367228,bolya2019yolact2,chen2019tensormask,chen2020blendmask,wang2020solo,10102514} predict the masks without extracting the object regions.
YOLACT~\cite{bolya2019yolact,bolya2019yolact2} predict candidate masks and then use anchors to predict the bounding box and mask weights for cropping and mixing masks to obtain the instance mask.
SOLO~\cite{wang2020solo} generates a dynamic convolution kernel for each instance and then adopts dynamic convolutions to estimate the instance mask.
Top-down methods have been extensively studied in the past. However, such methods usually require hand-crafted post-processing to refine the results and therefore are difficult to implement.

Bottom-up methods first extract features for each pixel and then generate the masks of objects by performing clustering.
SDLF~\cite{de2017semantic} trains the network with the metric learning loss to generate pixel feature representations and then uses mean-shift to cluster pixels with similar features to obtain instance segmentation.
%
%
SSAP~\cite{0SSAP} first uses CNN to extract features and then uses graph partition to cluster the pixels and generate instance masks.
Bottom-up methods cluster pixels based on the affinity between pixels, which are mostly designed for unsupervised instance segmentation and have poor performance on the supervised datasets.
Meanwhile, these methods need to process the pixel features to obtain the segmentation without category information, which lacks application value.

Direct methods extract features to directly predict the mask of each object.
QueryInst~\cite{fang2021queryinst} uses dynamic convolutions to predict the mask of instances on the basis of Sparse-RCNN~\cite{sun2021sparse}.
ISTR~\cite{wang2021end} uses a transformer encoder to refine features and then designs a mask decoder to predict instance segmentation.
SOLQ~\cite{dong2021solq} proposes a DCT mask decoder to predict instance segmentation.
MaskFormer~\cite{cheng2021maskformer} proposes to use the hidden state estimated to perform pixel-wise instance segmentation decoding.
Direct methods do not rely on non-maximum suppression to refine the predictions and are suitable for predicting the masks of adjacent human instances.
However, these methods usually use high-level features to predict the masks, which results in inaccurate object boundaries.

\begin{figure*}[!t]
\centering
\vspace{-2pt}
\includegraphics[width=0.95\linewidth]{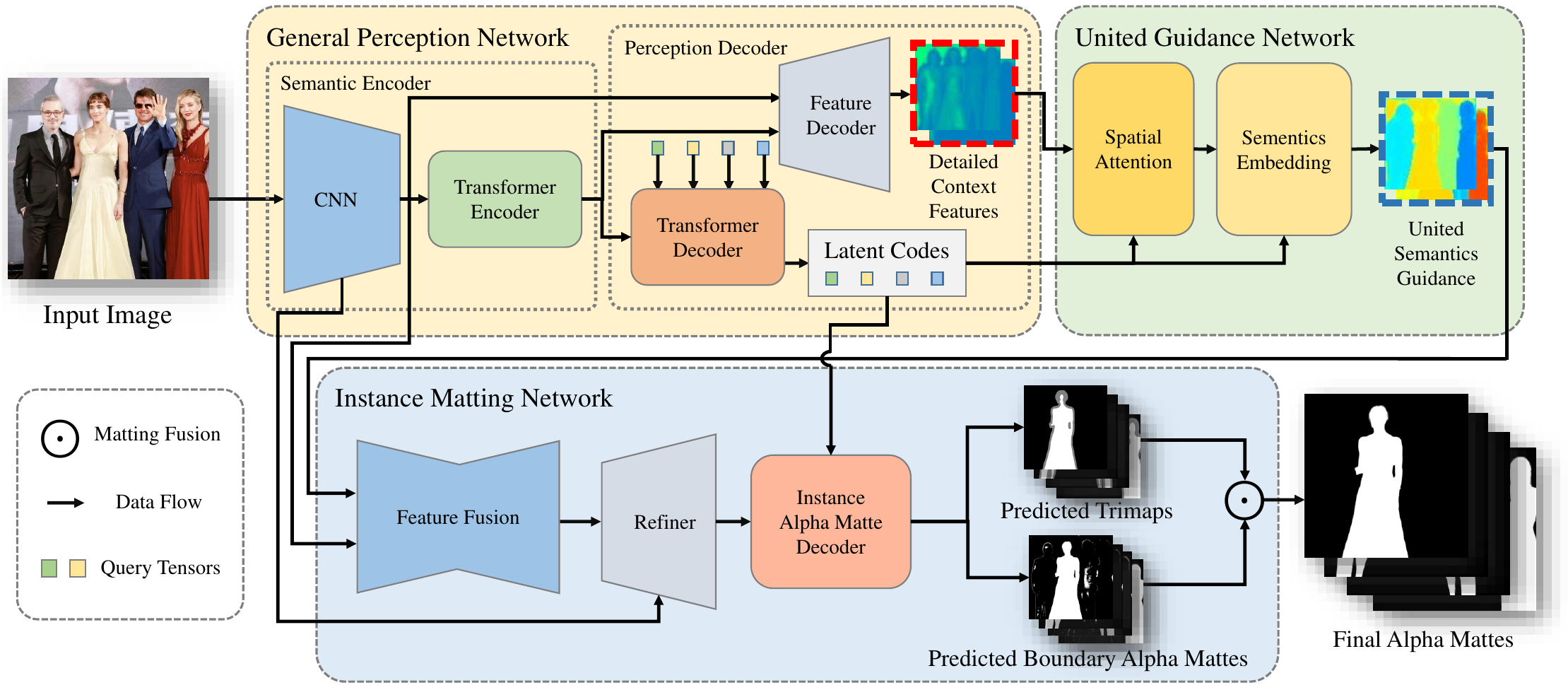}
\vspace{-5pt}
\caption{{Architecture of the proposed E2E-HIM.} The general perception network first extracts image features and instance latent codes. The united guidance network then utilizes the extracted image features and instance latent codes to generate the united semantics guidance. Finally, the instance matting network decodes the image features and united semantics guidance to estimate all instance-level alpha mattes.}
\vspace{-4pt}
\label{fig:overview}
\end{figure*}

\noindent\textbf{Image Matting.}
Image matting aims to predict the alpha matte of the foreground object.
There are mainly two types of image matting methods: trimap-based and automatic methods.

Trimap-based methods require a trimap indicating known foreground and background regions to estimate alpha mattes.
Currently, deep learning based methods~\cite{xu2017deep,2018AlphaGAN,tang2019learning,lu2019indices,forte2020fbamatting,liu2021lfpnet} have achieved good performance in this field.
DIM~\cite{xu2017deep} presents the first large-scale image matting dataset and for the first time uses the CNN to estimate the alpha matte with the trimap guidance.
With the image matting dataset, subsequent matting methods focus on improving the performance.
%
%
%
GCAMatting~\cite{li2020natural} uses the context module of the in-painting methods to integrate context information to improve the matting accuracy.
Deep Matte Prior~\cite{9634025} uses the image and matte priors to unsupervisedly train matting networks.
ATNet~\cite{9197694} proposes to use both the object-related details and the high-level semantics to estimate the alpha mattes.
FBAMatting~\cite{forte2020fbamatting} predicts the foreground, background, and alpha matte at the same time, which improves the accuracy of predicting the alpha matte.
MGMatting~\cite{yu2020mask} expands the kinds of auxiliary information, using coarse segmentation to help the network estimate the alpha matte.
LFPNet~\cite{liu2021lfpnet} uses long-range features to predict the alpha matte of high-resolution images.
ELGT-Matting~\cite{10011442} presents Global MSA and Window MSA to use wide correlations to estimate the alpha mattes.
Trimap-based methods can estimate the alpha matte of arbitrary objects, but the reliance on auxiliary inputs limits their application.

Automatic methods directly estimate the alpha matte of the foreground object in the input image without auxiliary inputs and have recently received much attention from researchers.
%
%
%
%
Hattmatting~\cite{2020Attention} uses hierarchical attention to aggregate semantic features to refine the boundaries.
MODNet~\cite{ke2020is} proposes a self-supervised strategy to optimize the network's generalization ability on real-world data.
%
%
P3MNet~\cite{2021Privacy} pays attention to privacy protection in the matting task and proposes a matting method that is robust to face occlusion.
%
%
SPAMattNet~\cite{10077601} adopts a two-stage network with dual-attention modules to first estimate the trimap and  then estimate the alpha matte.
RGB-D Human Matting~\cite{peng2023rgb} first estimates the coarse alpha matte from the depth map and then combines the RGB image to refine the predictions.
ISSMatting~\cite{instmattcrv} uses Mask R-CNN to generate instance-level trimap and then exploits DIM to estimate the alpha mattes.
InstMatt~\cite{sun2022instmatt} uses Mask R-CNN and MGMatting to estimate instance-level alpha mattes and use a refinement network to refine predictions.
Most automatic methods only estimate the alpha matte of all instances, which limits their applications in instance-level editing.

\section{Proposed Approach}
Given an input image $\boldsymbol{I}$ with multiple human instances, the proposed E2E-HIM aims to simultaneously predict all alpha mattes of individual human instances in the image.
To this end, we design an end-to-end human instance matting network including three sub-networks: a general perception network, a united guidance network, and an instance matting network as shown in Figure~\ref{fig:overview}.
Specifically, the general perception network first uses a semantic encoder to extract image features from $\boldsymbol{I}$, and then adopts a perception decoder to decode instance contexts into latent codes.
%
The united guidance network devises a spatial attention module and a semantics embedding module to generate the united semantics guidance that identifies the locations and semantic correspondences of the instances in the image.
The instance matting network fuse and refine the image features and united semantics guidance, then dynamically constructs decoders to predict all instance-level alpha mattes.

\begin{figure*}[!t]
\vspace{-2pt}
\centering
\resizebox{0.93\linewidth}{!} {
\includegraphics{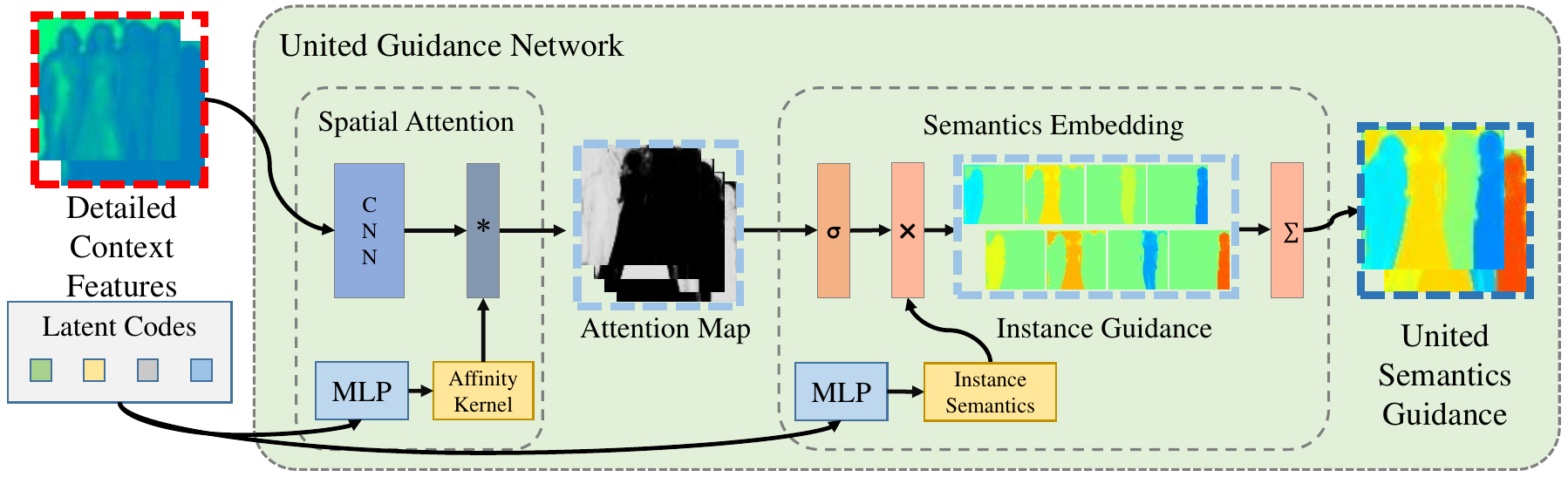}
}
\vspace{-5pt}
\caption{{Structure of the united guidance network.} A spatial attention module and a semantics embedding module are adopted to generate the guidance that identifies both the locations and semantic correspondences of the instances in the image.}
\vspace{-5pt}
\label{fig:ugn}
\end{figure*}

\subsection{General Perception Network}
\label{sec:31}
The general perception network aims to extract image features and then decode the instance contexts into latent codes.
%
To distinguish individual human instances, it is necessary to introduce detection or segmentation supervision.
However, previous object detection~\cite{girshickICCV15fastrcnn,ren2015faster,liu2016ssd,redmon2016you} and instance segmentation~\cite{he2020mask,bolya2019yolact,wang2020solo} frameworks predict many candidate instances and then use non-differentiable manual strategies to remove repetitive instances,
which are infeasible to be integrated into a united network for end-to-end training.
Recently, transformer-based detection and segmentation frameworks such as DETR~\cite{carion2020end} and MaskFormer~\cite{cheng2021maskformer} use matching-based supervision to  estimate the set of instances in a differentiable way, which facilitates the design of a united network for end-to-end training.
Inspired by these transformer-based methods, we formulate the decoding of instance latent codes as a set prediction problem and present a hybrid transformer network with matching-based instance segmentation supervision to solve the problem.

\noindent\textbf{Semantic Encoder.}
To extract semantic features for generating latent codes, we first adopt a CNN-transformer based semantic encoder.
Specifically, we adopt a deep stem to extract low-level image features from the input image $\boldsymbol{I}$.
To stabilize network training, we also use the group norm to improve the network performance under a small batch size.
A ResNet-50~\cite{he2016deep} backbone and a transformer encoder~\cite{vaswani2017attention}
are then used to extract high-level image features and perform context aggregation.
The context features  $\boldsymbol{F}_c\in\mathbb{R}^{C \times \frac{H}{16} \times \frac{W}{16}}$ of the transformer encoder are then processed by a perception decoder to generate the latent code for each human instance, where $C$ is the channel number of the feature maps.

\noindent\textbf{Perception Decoder.}
To decode the instance contexts into latent codes,  $\boldsymbol{F}_c$ is first flattened into a feature tensor $\boldsymbol{F}_{f}\in\mathbb{R}^{C \times \frac{H \times W}{256}}$.
Then, we define $N$ query features, where $N$ is larger than the instance number $X$, and use a $6$ layer transformer decoder network to extract instance contexts from the flattened features $\boldsymbol{F}_{f}$ and decode them into latent codes.
%
%
Given a learnable query tensor $\boldsymbol{Q} \in\mathbb{R}^{N \times C}$ that encodes $N$ query features, the multi-head self-attention of the transformer decoder first enhances the difference of each query feature.
%
%
Then, the multi-head cross-attention and feed-forward network of the transformer decoder use the affinity between the query features and the features $\boldsymbol{F}_{f}$ to extract the instance contexts and then decode them into a latent code $\boldsymbol{X}\in\mathbb{R}^{N \times C}$.

To train the general perception network, we introduce the auxiliary segmentation head and classification head (not shown in Figure~\ref{fig:overview}) to perform matching-based instance segmentation supervision.
Specifically, we use a multi-layer perception (MLP) to generate the segmentation convolution kernel $\boldsymbol{k}_{seg}\in\mathbb{R}^{N \times C \times 1 \times 1}$ and segmentation bias $\boldsymbol{b}_{seg}\in\mathbb{R}^{N \times 1 \times 1}$ from the latent code $\boldsymbol{X}$ as $\boldsymbol{k}_{seg} = \rm MLP (\boldsymbol{X})$ and $\boldsymbol{b}_{seg} = \rm MLP (\boldsymbol{X})$, where $\rm {MLP}$ is the MLP layer.
Then, we use the upsampling and convolution layers to build a feature decoder, which combines the image features from the CNN backbone and context features $\boldsymbol{F}_c$ to generate the detailed context features $\boldsymbol{F}_{dc}\in\mathbb{R}^{C\times \frac{H}{8} \times \frac{W}{8}}$.
Next, we use the segmentation convolution kernel $\boldsymbol{k}_{seg}$ with the segmentation bias $\boldsymbol{b}_{seg}$ to convolve detailed context features $\boldsymbol{F}_{dc}$ to obtain the human instance mask $\boldsymbol{M}_{pred}$
\begin{equation}
\boldsymbol{M}_{pred}={\rm {Conv}}( \boldsymbol{F}_{dc},\boldsymbol{k}_{seg})+{\rm{B}} (\boldsymbol{b}_{seg})
\end{equation}
where ${\rm{B}}(\cdot)$ is the broadcast function.
Additionly, we use MLPs to estimate the category $\boldsymbol{c}_{pred}\in\mathbb{R}^{N \times 2}$ (background or human instance) from the latent code $\boldsymbol{X}$.
Finally, we select ground-truth categories and masks that match the predicted categories and masks as labels to train the general perception network.

\subsection{United Guidance Network}
The united guidance network is to generate the united semantics guidance that facilitates  simultaneous instance-level matting.
As shown in Figure~\ref{fig:ugn}, the united guidance network consists of a spatial attention module and a semantics embedding module.
These components work together to encode the spatial locations and semantic correspondences of all the human instances in the image.

\noindent\textbf{Spatial Attention.}
\label{sec:afe}
To locate each human instance in the image,
we propose a spatial attention module to exploit the affinity between the context features and instance latent codes to automatically generate the instance-level homogeneous regions.
Specifically, we first adopt MLPs to generate the affinity kernel $\boldsymbol{k}_{a}\in\mathbb{R}^{N \times C\times1 \times 1}$ and affinity bias $\boldsymbol{b}_{a}\in\mathbb{R}^{N  \times1 \times 1}$ from the instance latent code $\boldsymbol{X}$.
Then, we use convolution layers to generate the spatial features $\boldsymbol{F}_{sp}\in\mathbb{R}^{C\times \frac{H}{8} \times \frac{W}{8}}$ and context-based background weight $\boldsymbol{W}_b\in\mathbb{R}^{1\times \frac{H}{8} \times \frac{W}{8}}$ from the detailed context features $\boldsymbol{F}_{dc}$ as $\boldsymbol{F}_{sp} = \rm CNN (\boldsymbol{F}_{dc})$ and $\boldsymbol{W}_b = \rm CNN (\boldsymbol{F}_{dc})$, where $\rm CNN$ is the convolutional layer.
Finally, the spatial attention map $\boldsymbol{W}_{sa}$ is obtained as
\begin{equation}
\boldsymbol{W}_{sa}={\rm {Concat}}(\boldsymbol{W}_b, {\rm {Conv}}( \boldsymbol{F}_{sp} , \boldsymbol{k}_{a} )+ {\rm{B}} (\boldsymbol{b}_{a}))
\label{eq:wsa}
\end{equation}
where ${\rm {Concat}}(\cdot)$ denotes the concatenate function.
%
Each channel of the spatial attention map represents the affinity of the pixels to the human instances or background.
To facilitate the subsequent processing, we reshape the attention map $\boldsymbol{W}_{sa}$ to a tensor with the shape of ${(N+1) \times 1 \times \frac{H}{8} \times \frac{W}{8}}$.

\begin{algorithm}[t]
\label{alg:genguidance}
\SetAlgoLined
\KwData{Detailed context features $\boldsymbol{F}_{dc}$,  and latent codes $\boldsymbol{X}$.}
\KwResult{United semantics guidance $\boldsymbol{G}_{All}$. }
Initialize $set \leftarrow \emptyset$\;
\For{$i \leftarrow 1$ \KwTo $S$}{
Compute the affinity kernel and affinity bias as $\boldsymbol{k}_{a} = \rm MLP (\boldsymbol{X})$,  $\boldsymbol{b}_{a}= \rm MLP (\boldsymbol{X})$\;
Compute the spatial features and context-based background weight as $\boldsymbol{F}_{sp}  = \rm CNN (\boldsymbol{F}_{dc})$, $\boldsymbol{W}_b = \rm CNN (\boldsymbol{F}_{dc})$\;
Compute the spatial attention map as $\boldsymbol{W}_{sa}={\rm {Concat}}(\boldsymbol{W}_b, {\rm {Conv}}( \boldsymbol{F}_{sp} , \boldsymbol{k}_{a} )+ {\rm{B}} (\boldsymbol{b}_{a}))$\;
Compute the instance semantics features as $\boldsymbol{F}_{rep} = \rm MLP  (\boldsymbol{X})$\;
Upsample the instance semantics features as $\boldsymbol{F}_{urep} =  \rm  Upsample (\boldsymbol{F}_{rep})$\;
Compute the context-based background features as $\boldsymbol{F}_{bf} = \rm CNN (\boldsymbol{F}_{dc})$\;
Compute the guidance $\boldsymbol{G}^i={\rm {Sum}}({\rm {Softmax}}(\boldsymbol{W}_{sa}, 1)  \odot {\rm {Concat}}(\boldsymbol{F}_{bf} , \boldsymbol{F}_{urep}), 1)$\;
$set  \leftarrow set \cup \boldsymbol{G}^i$\;
}
Generate the united semantics guidance as $\boldsymbol{G}_{All}={\rm {Concat}}({\boldsymbol{G}^1}, {\boldsymbol{G}^2}, ... {\boldsymbol{G}^S})$\;
\caption{Guidance generation procedure.}
\end{algorithm}

\noindent\textbf{Semantics Embedding.}
To generate the united semantics guidance that identifies both the locations and semantic correspondences of all instances for alpha matte estimation, we propose the semantics embedding module to embed the instance semantics that is generated from the latent codes into the spatial attention map.
%
Inspired by the non-overlapping nature of the regions of instances, we embed the instance semantics by modulating the instance semantics with the spatial attention map.
%
Specifically, we first use the softmax activation to sparsen the spatial attention map, which makes each location associated with the most relevant instance.
Then, we use MLPs to generate the instance semantics features $\boldsymbol{F}_{rep}\in\mathbb{R}^{N \times C}$ from the latent code $\boldsymbol{X}$.
We upsample $\boldsymbol{F}_{rep}$ to the same shape of $\boldsymbol{F}_{dc}$ and generate the upsampled instance semantics features $\boldsymbol{F}_{urep}\in\mathbb{R}^{N \times C\times \frac{H}{8} \times \frac{W}{8}}$.
Next, we use convolution layers to generate the context-based background features $\boldsymbol{F}_{bf}\in\mathbb{R}^{1 \times C\times \frac{H}{8} \times \frac{W}{8}}$ from the detailed context features $\boldsymbol{F}_{dc}$ as  $\boldsymbol{F}_{bf} = \rm CNN (\boldsymbol{F}_{dc})$.
Finally, we use the attention map $\boldsymbol{W}_{sa}$ to modulate the instance semantics $\boldsymbol{F}_{rep}$ and features  $\boldsymbol{F}_{bf}$ to embed the instance semantics into the untied semantics guidance $\boldsymbol{G}$ as
\begin{equation}
\boldsymbol{G}={\rm {Sum}}({\rm {Softmax}}(\boldsymbol{W}_{sa}, 1)  \odot {\rm {Concat}}(\boldsymbol{F}_{bf} , \boldsymbol{F}_{urep}), 1)
\label{eq:guidance}
\end{equation}
where ${\rm {Softmax}}(\boldsymbol{T}, i)$ denotes the softmax activation function computed along the $i$-th dimension of the input tensor $\boldsymbol{T}$.
${\rm {Sum}}(\boldsymbol{T}, i)$ denotes the function that computes the sum of elements across the $i$-th dimension of the  tensor $\boldsymbol{T}$.
%
The embedded instance semantics indicates the regions corresponding to each instance, which helps to guide instance-level matting.

\begin{figure}[!t]
\centering{
        \includegraphics[width=0.48\textwidth]{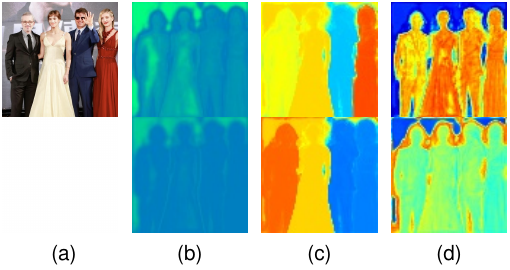}
}
\caption{Visualization of the learned guidance. (a) Input image. (b) Context features from the General Perception Network. (c) Features of the first guidance head. (d) Features of the second guidance head. }
\label{fig:gd}
\end{figure}

\noindent\textbf{Guidance Generation.}
To generate the guidances that focus on different regions of each instance, we adopt a multi-head guidance design as described in Algorithm~\ref{alg:genguidance}.
Specifically, we define the combination of a spatial attention module and a semantics embedding module as a guidance head.
Then, we build the united guidance network with ${S}$ guidance heads to generate ${S}$ temporary guidances ${\boldsymbol{G}^1}, {\boldsymbol{G}^2}, ... {\boldsymbol{G}^S}$.
Finally, the united semantics guidance $\boldsymbol{G}_{All}$ is obtained by concatenating all temporary guidances as
\begin{equation}
\boldsymbol{G}_{All}={\rm {Concat}}({\boldsymbol{G}^1}, {\boldsymbol{G}^2}, ... {\boldsymbol{G}^S})
\label{eq:gall}
\end{equation}
To facilitate the understanding of the united guidance network, we provide visualizations of the guidances generated from the context features in Figure~\ref{fig:gd}.
As evident from the visualizations, the features of two guidance heads exhibit improved discrimination between human-human and human-background regions compared to the original context features.
Specifically, the combination of the features of the first guidance head  helps in discriminating different human instances, facilitating the estimation of human instance contours.
The combination of the features of the second guidance head   helps in discriminating human regions, background regions, and boundary regions, improving the estimation of alpha mattes at the boundaries.



\subsection{Instance Matting Network}
The instance matting network aims to decode the image features and the united semantics guidance to simultaneously estimate all instance-level alpha mattes.
As shown in Figure~\ref{fig:overview}, we first propose a feature fusion and a refiner to aggregate the features for each instance and recover the low-level image features.
Then, we utilize the latent codes to dynamically construct the instance alpha matte decoder to estimate instance alpha mattes.
Due to the highly imbalanced distribution of foreground regions, background regions, and unknown regions (boundary), directly training the network to predict alpha matte fails to converge.
%
Therefore, we construct two decoders to independently estimate the instance-level trimaps and boundary alpha mattes, which stabilizes the network training.

\noindent\textbf{Feature Fusion and Refiner.}
The image features from the CNN backbone of the general perception network are not well-aggregated and lack low-level image features, which are not feasible for alpha matte estimation.
To aggregate the features for each instance and recover the low-level image features, we adopt fully convolutional networks to construct the feature fusion and the refiner for feature processing.
Specifically, we first concatenate the image features and the united semantics guidance to combine the image features with the locations and semantic correspondences of human instances.
Then, we use a simple U-Net to fuse the information, which utilizes the guidance to aggregate the image features and generate the instance-aware image features.
Next, we use upsampling convolution layers to refine the instance-aware image features with low-level image features from the stem of the general perception network, which helps to estimate accurate trimaps and alpha mattes at the boundaries.
Finally, we use the refined instance-aware image features to generate the alpha matte features $\boldsymbol{F}_\alpha \in\mathbb{R}^{C_\alpha \times {H} \times {W}}$ for  alpha matte decoding and the trimap features $\boldsymbol{F}_{tri} \in\mathbb{R}^{C_{tri} \times \frac{H}{2} \times \frac{W}{2}}$ for trimap decoding, where $C_{\alpha}$ and $C_{tri}$ are the feature channel numbers.

\noindent\textbf{Instance Alpha Matte Decoder.}
With the alpha matte features $\boldsymbol{F}_\alpha$ and trimap features $\boldsymbol{F}_{tri}$, we construct decoders to predict instance-level alpha mattes in parallel.
Since the distribution of the foreground regions, background regions, and unknown regions of the alpha mattes are imbalanced, training the network to directly estimate the alpha mattes causes overly smooth estimation in the unknown regions.
To tackle this problem, we construct two kinds of decoders to estimate the instance-level trimaps and boundary alpha mattes.

To estimate the instance trimaps and alpha mattes, we use the latent codes to dynamically build instance-level decoders for prediction.
Specifically, to estimate the trimaps, we use MLPs to generate the trimap decode kernel $\boldsymbol{k}_{tri} \in\mathbb{R}^{(N \times 3) \times C_{tri}  \times 1 \times 1}$ and bias $\boldsymbol{b}_{tri} \in\mathbb{R}^{(N \times 3) \times1 \times 1}$ from the latent code $\boldsymbol{X}$.
We use the trimap kernel $\boldsymbol{k}_{tri}$ and trimap bias $\boldsymbol{b}_{tri}$ to build convolutions to process the trimap features $\boldsymbol{F}_{tri}$ and obtain the predicted trimap $\boldsymbol{T}_{pred} \in\mathbb{R}^{(N \times 3) \times \frac{H}{2} \times \frac{W}{2}}$ as
\begin{equation}
\boldsymbol{T}_{pred}={\rm {Conv}}( \boldsymbol{F}_{tri},\boldsymbol{k}_{tri})+{\rm{B}}(\boldsymbol{b}_{tri})
\end{equation}
To facilitate the fusion with the alpha mattes, we upsample and reshape the predicted trimap $\boldsymbol{T}_{pred}$ to a tensor with  the shape of $N \times 3 \times H \times W$.
To estimate the alpha matte at the unknown regions of each human instance, we use MLPs to generate the alpha matte kernel $\boldsymbol{k}_\alpha \in\mathbb{R}^{N \times C_\alpha  \times 1 \times 1}$ and alpha matte bias $\boldsymbol{b}_\alpha \in\mathbb{R}^{N  \times1 \times 1}$ from the latent code $\boldsymbol{X}$.
We use the alpha matte kernel $\boldsymbol{k}_\alpha$ and alpha matte bias $\boldsymbol{b}_\alpha$ to build convolutions to process the alpha matte features $\boldsymbol{F}_\alpha$ and obtain the predicted boundary alpha matte tensor $\boldsymbol{\alpha}_{pred} \in\mathbb{R}^{N  \times {H} \times {W}}$ as
\begin{equation}
\boldsymbol{\alpha}_{pred}={\rm {Conv}}( \boldsymbol{F}_{\alpha},\boldsymbol{k}_{\alpha})+{\rm{B}}(\boldsymbol{b}_{\alpha})
\end{equation}
We further reshape the predicted boundary alpha matte $\boldsymbol{\alpha}_{pred}$ to a tensor with the shape of $N \times 1 \times H \times W$.
The final alpha matte tensor $\boldsymbol{\alpha}_{fin}$ is obtained  through the matting fusion processing of trimap-based matting methods~\cite{xu2017deep,li2020natural} as
\begin{equation}
\boldsymbol{\alpha}_{fin}=\boldsymbol{\alpha}_{pred}\odot \boldsymbol{U}_{pred}+\boldsymbol{F}_{pred}
\label{eq:fin}
\end{equation}
where $\boldsymbol{U}_{pred}$ and $\boldsymbol{F}_{pred}$ are the unknown regions and foreground regions in the estimated trimap, respectively.

\subsection{Loss Function}
To train the proposed network, we construct loss functions for the outputs of the general perception network and the instance matting network.
Specifically, we first follow MaskFormer~\cite{cheng2021maskformer} to obtain the sorted ground truth labels corresponding to the latent codes with minimum-cost bipartite matching.
The matching cost $\mathcal{C}_{match}$ for the $i$-th latent code to the $x$-th instance is composed of the class matching cost $\mathcal{C}_{mcls}$, focal segmentation matching cost $\mathcal{C}_{mfocal}$, and dice segmentation matching cost $\mathcal{C}_{mdice}$ as
\begin{equation}
\mathcal{C}_{match}^{i,x}=\mathcal{C}_{mcls}^{i}+  \mathcal{C}_{mfocal}^{i,x}+  \mathcal{C}_{mdice}^{i,x}
\label{eq:lmatch}
\end{equation}
The class matching cost $\mathcal{C}_{mcls}$ measures the negative probability of classifying a latent code as a human instance, which is defined as
\begin{equation}
\mathcal{C}_{mcls}^{i}= -\frac{\exp{({c}^{i,2}_{pred})}}{\exp{({c}^{i,1}_{pred})}+\exp{({c}^{i,2}_{pred})}}
\end{equation}
where $\boldsymbol{c}_{pred}$ is the  categories predicted by the perception decoder.
The focal segmentation matching cost $\mathcal{L}_{mfocal}$ measures the focal cross entropy between the predicted instance masks $\boldsymbol{M}_{pred}$ and ground truth instance masks $\boldsymbol{M}_{gt}$, which is defined as
\begin{equation}
\mathcal{C}_{mfocal}^{i,x}={\rm{{BCE}_{focal}}}(M_{pred}^i,M^x_{gt})
\end{equation}
where $\rm{{BCE}_{focal}}(\cdot)$ denotes the focal binary cross entropy function~\cite{lin2020focal}.
The dice segmentation matching cost $\mathcal{L}_{mdice}$ measures the dice similarity coefficient between the predicted  instance masks $\boldsymbol{M}_{pred}$ and ground truth instance masks $\boldsymbol{M}_{gt}$, which is defined as
\begin{equation}
\mathcal{C}_{mdice}^{i,x}={\rm{{Dice}}}(M_{pred}^i,M^x_{gt})
\end{equation}
where $\rm{Dice}(\cdot)$ denotes the dice function~\cite{milletari2016v}.
With the matching costs, we adopt 2D rectangular assignment to obtain the instance indexes corresponding to the latent codes and then generate the sorted ground truth categories $\boldsymbol{c}_{sort}$, masks $\boldsymbol{M}_{sort}$, trimaps $\boldsymbol{T}_{sort}$, and alpha mattes $\boldsymbol{\alpha}_{sort}$ according to the ground truth labels.

After obtaining the sorted ground truths, we construct the loss functions used to train the network.
Specifically, to supervise the general perception network, we construct the perception loss as
\begin{equation}
\mathcal{L}_{p}=\lambda_{c} \mathcal{L}_{cls}+ \lambda_{s} \mathcal{L}_{seg}
\end{equation}
where $\mathcal{L}_{cls}$ and $\mathcal{L}_{seg}$  are the classification loss and segmentation loss, respectively.
$\lambda_{c}$ and $\lambda_{s}$ are the weights to balance the two losses.
The classification loss $\mathcal{L}_{cls}$  measures the cross entropy between the predicted categories $\boldsymbol{c}_{pred}$ and sorted ground truth categories $\boldsymbol{c}_{sort}$, which is defined as
\begin{equation}
\mathcal{L}_{cls}=\frac{1}{N} \sum_{i=1}^{N}{w^{i}  {\rm{CE}}(c_{pred}^i,c^i_{sort})}
\end{equation}
where ${\rm{CE}}(\cdot)$ denotes the cross entropy function.
%
%
$w^{i}$ is the category weight.
The segmentation loss $\mathcal{L}_{seg}$ measures the focal binary cross entropy and dice similarity coefficient between the predicted instance masks $\boldsymbol{M}_{pred}$ and sorted ground truth instance masks $\boldsymbol{M}_{sort}$, which is defined as
\begin{equation}
\begin{split}
\mathcal{L}_{seg}=&\frac{1}{|\Omega|} \sum_{i\in \Omega}{\lambda_{b}\rm{{BCE}_{focal}}(M_{pred}^i,M^i_{sort})}\\
&+\frac{1}{|\Omega|} \sum_{i\in \Omega}\lambda_{d}{\rm{Dice}}(M_{pred}^i,M^i_{sort})
\end{split}
\end{equation}
where $\Omega$ denotes the set of the indexes of the latent codes that have the corresponding human instances.
$\lambda_{b}$ and $\lambda_{d}$ are the weights to balance the two losses.
To supervise the network to estimate the trimap and alpha matte, we define the matting loss as
\begin{equation}
\mathcal{L}_{m}=\lambda_{t} \mathcal{L}_{tri}+ \lambda_{\alpha} \mathcal{L}_{\alpha}
\end{equation}
where $\mathcal{L}_{tri}$ and $\mathcal{L}_{\alpha}$ are the trimap loss and alpha matte loss, respectively.
%
$\lambda_{t}$ and $\lambda_{\alpha}$ are the weights to balance the two losses.
The trimap loss $\mathcal{L}_{tri}$ measures the focal cross entropy between the predicted trimaps $\boldsymbol{T}_{pred}$ and sorted ground truth trimaps $\boldsymbol{T}_{sort}$, which is defined as
\begin{equation}
\mathcal{L}_{tri}= \frac{1}{|\Omega|} \sum_{i\in \Omega} {\rm{{CE}_{focal}}}(T_{pred}^i,T_{sort}^i)
\end{equation}
The alpha matte loss $\mathcal{L}_{\alpha}$ measures the mean absolute errors between the predicted boundary alpha mattes $\boldsymbol{\alpha}_{pred}$ and sorted ground truth alpha mattes $\boldsymbol{\alpha}_{sort}$, which is defined as
\begin{equation}
\begin{split}
\mathcal{L}_{\alpha}=&\lambda_{pu}\frac{1}{\sum_{i\in \Omega}| U^i_{pred}| } \sum_{i\in \Omega} \sum_{j\in U^i_{pred}}  |\alpha_{pred}^{i,j}-\alpha_{sort}^{i,j}|\\
&+\lambda_{gu}\frac{1}{\sum_{i\in \Omega} | U^i_{sort}| } \sum_{i\in \Omega}  \sum_{j\in U^i_{sort}}  |\alpha_{pred}^{i,j}-\alpha_{sort}^{i,j} |
\end{split}
\end{equation}
where $\lambda_{pu}$ and $\lambda_{gu}$ are the weights to balance the two losses.
$\alpha_{pred}^{i,j}$ and$\alpha_{sort}^{i,j}$ are the predicted boundary alpha mattes and ground truth alpha mattes for the latent code $\boldsymbol{X}^i$ in pixel $j$.
$U^i_{pred}$ and $U^i_{sort}$ are the predicted and ground truth unknown regions for the latent code $\boldsymbol{X}^i$.
The overall loss of the network is defined as
\begin{equation}
\mathcal{L}= \lambda_{p} \mathcal{L}_{p}+\lambda_{m} \mathcal{L}_{m}
\label{eq:lossall}
\end{equation}
where $\lambda_{p}$ and $\lambda_{m}$ are the weights to balance the two losses.

\begin{figure*}[!t]
\begin{center}
\subfloat[Synthetic human images]{
\includegraphics[height=7.75cm]{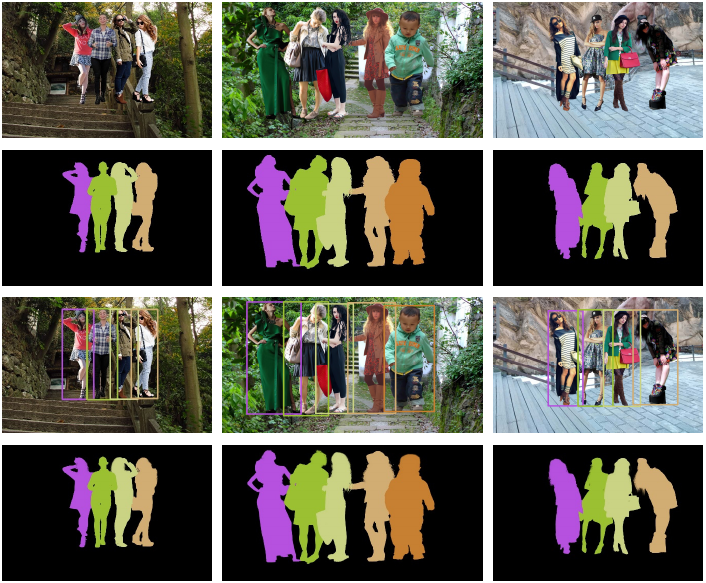}
}
\hfill
\subfloat[Real-world human images]{
\includegraphics[height=7.75cm]{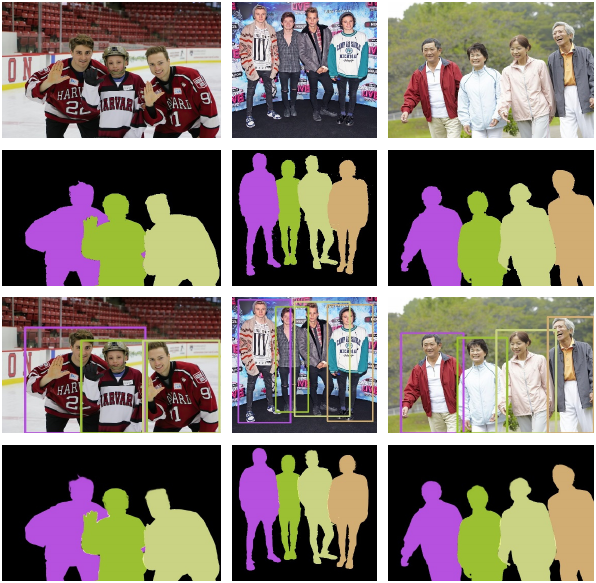}
}
\end{center}
\vspace{-5pt}
\caption{{Examples of images in the proposed HIM-100K dataset. } From top to bottom in the figure are the original images, instance segmentation annotations, instance bounding box annotations, and instance alpha matte annotations. The human instances in the images are marked in different colors.}
\label{fig:himexp}
\end{figure*}

\section{Experiments}
In this section, we evaluate the performance of the proposed E2E-HIM\footnote{The demo code is available at \url{https://github.com/QLYoo/E2E-HIM}.} on both our collected HIM-100K dataset and three traditional human matting datasets.

\subsection{Datasets}
\subsubsection{HIM-100K}
To train and evaluate human instance matting methods, we construct a large-scale human instance matting dataset (HIM-100K).
Specifically, we first collect real-world human images from a variety of sources. 
These sources include existing datasets such as MHP~\cite{mhp2018}, Dinstinctions-646~\cite{2020Attention}, and supervisely~\cite{supervisely}, each adhering to their respective license constraints.
Additionally, we collect human images from the internet  (e.g., pexels~\cite{pexels}), which are not permissible for commercial use.
Furthermore, we purchase human images from a commercial portrait photographer, which are authorized for commercial application and distribution.
To avoid ethical issues, we meticulously  filter out any images that may contain violent or inappropriate content.
Then, we collect human-free images from the Internet as background images.
Next, we generate the real-world data of HIM-100K by enlisting image editors to annotate the images using Photoshop with commercial matting plugins. 
Finally,  we generate the synthetic data of HIM-100K by composting the annotated real-world human images with the collected background images.

To facilitate related research, each human instance in an image is annotated with three labels including the alpha matte, bounding box, and segmentation.
Specifically, HIM-100K contains 47,980 real-world human images and 95,597 synthetic human images.
The real-world human images vary from 1,024 to 1,280 on the long side.
The synthetic human images vary from 1,000 to 2,000 on the long side.
The training set of HIM-100K consists of 44,980 real-world human images and 95,597 synthetic human images, each of which contains 1 to 12 human instances.
There are a total of 326,455 human instances in the training set.
The validation set of HIM-100K consists of 3,200 real-world human images, each of which contains 1 to 6 human instances.
There are a total of 8,624 human instances in the validation set.
As shown in Figure~\ref{fig:himexp}, we give examples of the synthetic and real-world human images in the HIM-100K dataset.
The HIM-100K dataset will be made available to researchers for academic use.

\subsubsection{PPM-100}
The Photographic Portrait Matting dataset (PPM-100) \cite{ke2020is} is a human matting validation set, which collects 100 real-world portrait images from multiple real-world scenarios.
The images in the PPM-100 dataset have disturbances such as blur, decorations, and richer postures, which pose a great challenge for human matting methods.
We use the PPM-100 dataset to evaluate the generalization ability of E2E-HIM trained on HIM-100K.

\subsubsection{RWP-636}
The Real-world Portrait dataset (RWP-636) \cite{zhang2020mask} is a human matting validation set, which collects 636 real-world high-resolution portrait images.
To ensure that the dataset can evaluate the performance of methods in real-world scenarios, the images in the RWP-636 dataset are of different image qualities and the humans in the images are in diverse poses.
We use the RWP-636 dataset to evaluate the generalization ability of E2E-HIM trained on HIM-100K.

\subsubsection{P3M}
The Privacy-Preserving Portrait Matting dataset (P3M) \cite{2021Privacy} is a human matting dataset, which takes into account the protection of facial privacy.
The training set consists of 9,421 portrait images where the faces are obscured.
The validation set comprises two subsets: P3M-500-P and P3M-500-NP, each containing 500 portrait images with obscured faces and 500 normal portrait images, respectively.
We follow P3MNet~\cite{2021Privacy} to use the P3M dataset for training and evaluating E2E-HIM. In addition, we use the P3M-500-NP validation set to evaluate the generalization ability of E2E-HIM trained on HIM-100K.

\subsection{Implementation Details}
We implement the proposed E2E-HIM using the PyTorch~\cite{NEURIPS2019_9015} framework.
Specifically, we adopt a united guidance network with two heads and 20 queries for E2E-HIM.
Our E2E-HIM is trained on four NVIDIA RTX 2080Ti GPUs with a total batch size of 4 (1 per GPU) for 40 epochs.
During the data preprocessing stage, we first randomly read an image from the HIM-100K dataset.
Then, we follow image matting methods~\cite{li2020natural,forte2020fbamatting} to randomly composite human instances in a random image of HIM-100K onto the current image.
Next,  we perform data augmentation on the current image by applying random horizontal flipping, random affine transform, random contrast transform, random gamma transform, and random saturation transform.
Finally, the current image is randomly cropped to a $640 \times 640$ image patch and fed to the network.
To accelerate training the network, we initialize the Resnet-50~\cite{he2016deep} backbone with the weights pre-trained on ImageNet~\cite{deng2009imagenet}.
The other parameters are initialized with the Kaiming initializer~\cite{he2015delving}.
To optimize the network weights, we adopt the AdamW optimizer~\cite{loshchilov2018decoupled} with the betas of (0.9, 0.999) and a weight decay of 0.0005.
The optimizer is initialized with a learning rate of 0.00008 and then adjusted using a cosine annealing schedule.
We set the coefficients in the loss functions as $\lambda_{c}=5.$, $\lambda_{s}=1.$, $\lambda_{b}=0.1$, $\lambda_{d}=1$, $\lambda_{t}=10$, $\lambda_{\alpha}=5$, $\lambda_{pu}=3$, $\lambda_{gu}=5$, $\lambda_{p}=1$, $\lambda_{m}=1$, $w^i=0.1$ for the background category, and $w^i=1$ for the human category.

\subsection{Metrics}
To evaluate the  performance of the human instance matting methods, we propose four new metrics: ACC (accuracy), REC (recall), EMSE (effective mean square error), and EMAD (effective mean absolute difference), which are defined as
\begin{equation}
\begin{split}
{\rm{ACC}_{TH}}&=\frac{N_{TH}}{N_{pred}} \\
{\rm{REC}_{TH}}&=\frac{N_{TH}}{N_{gt}}\\
{\rm{EMSE}_{TH}}&=\frac{1}{N_{TH}} \sum_{i \in \Omega_{TH}} (\frac{1}{|{\Pi_i}|} \sum_{j \in {\Pi}_i}{({\alpha^{i,j}_{pred}}-{\alpha^{i,j}_{gt}})^2}) \\
{\rm{EMAD}_{TH}}&=\frac{1}{N_{TH}} \sum_{i \in \Omega_{TH}}  (\frac{1}{|{\Pi_i}|} \sum_{j \in {\Pi}_i}  |{\alpha^{i,j}_{pred}}-{\alpha^{i,j}_{gt}}|) \\
\end{split}
\end{equation}
where $N_{TH}$ is the number of the predicted alpha mattes with IoU higher than a threshold $TH$.
$N_{pred}$ and $N_{gt}$ are the numbers of the predicted alpha mattes and ground truth alpha mattes, respectively.
$\Omega_{TH}$ is the set of indexes of the predicted alpha mattes with IoU higher than a threshold $TH$.
${\Pi}_i$ is the set of all pixels in the $i$-th alpha matte.
$\alpha_{pred}^{i,j}$ and $\alpha^{i,j}_{gt}$ are the $i$-th predicted and ground truth alpha mattes at pixel $j$.
The ACC metric evaluates the ratio of correct predictions to the total predictions.
The REC metric evaluates the ratio of correct predictions to the ground truths.
The EMSE and EMAD metrics evaluate the errors between the correctly predicted alpha mattes and ground truth alpha mattes.
%
In this paper, we adopt the IoU thresholds $TH$ of 0.5 and 0.75 for evaluation.
To evaluate the performance of human matting, we follow MODNet~\cite{ke2020is} and P3MNet~\cite{2021Privacy} to adopt SAD (sum of absolute differences), MSE (mean square error), and MAD (mean absolute difference) as metrics to evaluate the errors between the predictions and the ground truths.

\begin{table*}[!t]
\centering
\caption{
Quantitative results of human instance matting on the HIM-100K dataset.  $*$ and $\dag$ indicate the instance segmentation methods trained on the MS-COCO~\cite{lin2014microsoft} and our HIM-100K datasets, respectively. }
\resizebox{\textwidth}{!}{
\begin{tabular}{c|c|c|c|c|c|c|c|c|c|c}
\toprule
Method&Segmentation & Matting & $\rm{EMSE}_{0.5}$ & $\rm{EMAD}_{0.5}$ & $\rm{REC}_{0.5}$ & $\rm{ACC}_{0.5}$ & $\rm{EMSE}_{0.75}$ & $\rm{EMAD}_{0.75}$ & $\rm{REC}_{0.75}$ & $\rm{ACC}_{0.75}$ \\
\midrule
ISSMatting&Mask R-CNN$^*$   & DIM     &  0.0193&0.0215&0.9526&0.7844&0.0128&0.0148&0.8328&0.6858 \\
InstMatt&Mask R-CNN$^*$   & FBAMatting     & 0.0187  & 0.0209  & 0.9549  & 0.7863  & 0.0122  & 0.0142  & 0.8422  & 0.6935  \\
InstMatt&Mask R-CNN$^*$   & MGMatting    & 0.0259 & 	0.0311& 0.7517& 0.6190&  0.0131 &	0.0170& 	0.5531& 0.4554
\\
\midrule
&Mask R-CNN$^\dag$   & FBAMatting     & 0.0171  & 0.0192  & 0.9485  & 0.8446  & 0.0104  & 0.0123  & 0.8367  & 0.7451  \\
&Mask R-CNN$^\dag$   & MGMatting    & 0.0237  &	0.0288  &	0.7879  &	0.7016  &	0.0123  &	0.0161 &	0.5990 &	0.5334
\\
&PointRend$^\dag$   & FBAMatting     & 0.0138  & 0.0158  & 0.9632  & 0.8809  & 0.0099  & 0.0118  & 0.8845  & 0.8089  \\
&PointRend$^\dag$   & MGMatting    & 0.0202&	0.0249&	0.8508&	0.7781&	0.0110&	0.0146&	0.6939&	0.6346
\\
Combination  &SOLO$^\dag$   & FBAMatting     & 0.0122  & 0.0140  & 0.9276  & \textbf{0.9873}  & 0.0095  & 0.0118  & 0.8780  & 0.9345  \\
Schemes   &SOLO$^\dag$   & MGMatting    & 0.0201&	0.0251&			0.7846&	0.8351&	0.0120&	0.0158&			0.6369&	0.6778
\\
&QueryInst$^\dag$   & FBAMatting     & 0.0102  & 0.0120  & 0.9199  & 0.9834  & 0.0078  & 0.0096  & 0.8799  & \textbf{0.9406}  \\
&QueryInst$^\dag$   & MGMatting    & 0.0195& 	0.0244&			0.7823&	0.8363&	0.0111& 	0.0149&		0.6355&	0.6794
\\
&MaskFormer$^\dag$ & FBAMatting     & 0.0107&	0.0122&	0.9760&	0.9308&0.0084&	0.0099	&	0.9111&0.8688  \\
&MaskFormer$^\dag$ & MGMatting    & 0.0142	&0.0182	 & 0.8973& 0.8557 	&0.0093	&0.0127 &	0.7795&	0.7433 \\
\midrule
E2E-HIM & -&-     & \textbf{0.0076}  & \textbf{0.0090}  & \textbf{0.9856}  & 0.9658  & \textbf{0.0060}  &\textbf{0.0074}  &\textbf{0.9486}  & 0.9296  \\
\bottomrule
\end{tabular}%
}
\label{tab:him100k}%
\end{table*}%

\begin{figure*}[!th]
\centering{
\includegraphics[width=1.\linewidth]{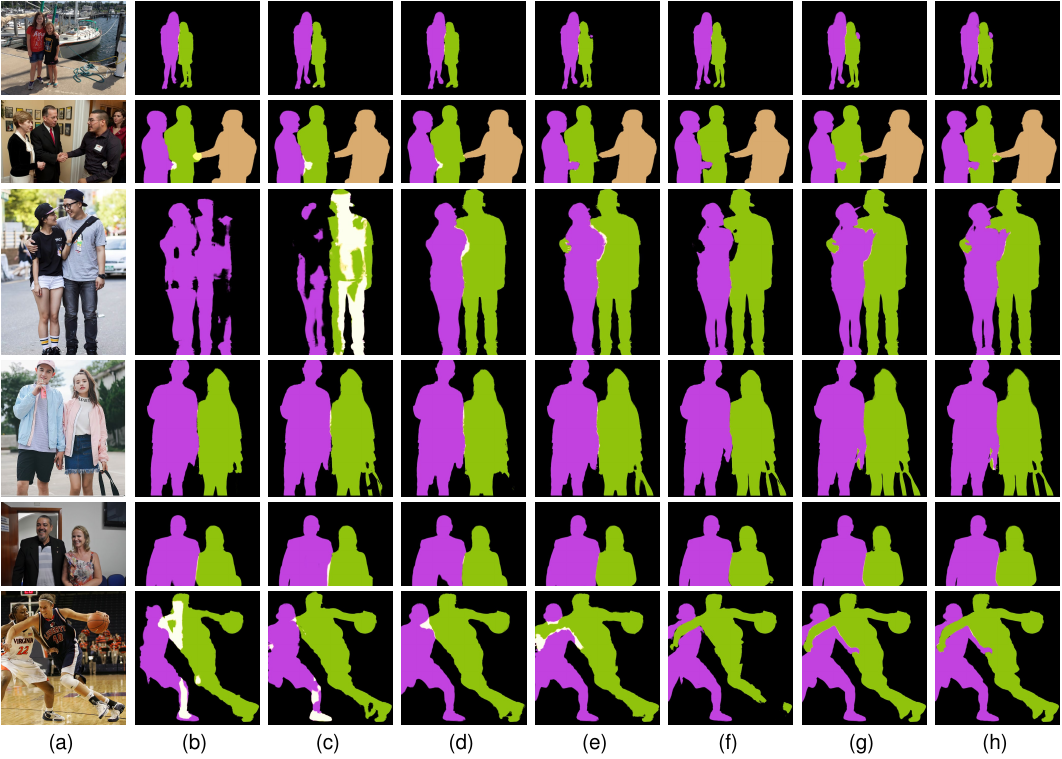}
}
\caption{{Qualitative results of human instance matting on the HIM-100K dataset.} The human instances in each image are marked in different colors.  (a) Input image. (b) Mask R-CNN + FBAMatting. (c) PointRend + FBAMatting. (d) SOLO + FBAMatting. (e) QueryInst + FBAMatting. (f) MaskFormer + FBAMatting. (g) E2E-HIM. (h) Ground Truth. Zoom in for the best visualization.}
\label{fig:him100}
\end{figure*}

\subsection{Results on Human Instance Matting}
\label{sec:him100}
To validate the effectiveness of the proposed method on human instance matting, we select two state-of-the-art human instance matting methods including: ISSMatting~\cite{instmattcrv} (Mask R-CNN~\cite{he2020mask} + DIM~\cite{xu2017deep}) and InstMatt~\cite{sun2022instmatt} (Mask R-CNN + FBAMatting~\cite{forte2020fbamatting}, and Mask R-CNN + MGMatting~\cite{yu2020mask}).
In particular, the Mask R-CNN used by these methods~\cite{instmattcrv,sun2022instmatt} is trained on the  MS-COCO~\cite{lin2014microsoft} dataset as described in their papers.
In addition, we implement several combinations of exiting instance segmentation methods (Mask R-CNN, PointRend~\cite{kirillov2020pointrend}, SOLO~\cite{wang2020solo}, QueryInst~\cite{fang2021instances}, and MaskFormer~\cite{cheng2021maskformer}) and matting methods (FBAMatting and MGMatting~\cite{yu2020mask}).
The instance segmentation methods used by the combinations is trained on the proposed HIM-100K dataset.
We evaluate these methods and summarize the quantitative results in Table~\ref{tab:him100k}.
Based on the results, we have two empirical findings:
\emph{First}, the matting methods (FBAMatting and MGMatting) have a significant impact on the alpha matte accuracy of the instance segmentation-based methods.
FBAMatting significantly reduces the EMSE and EMAD, which indicates that FBAmatting can estimate the high-quality alpha mattes with the instance segmentation.
However, due to domain drift such as the scale difference between the HIM-100K dataset and the training set of MGMatting (Adobe Composite-1K~\cite{xu2017deep}), MGMatting performs poorly and has higher EMSE and EMAD.
%
\emph{Second}, E2E-HIM has low EMAD and EMSE, and high REC, which indicates that our methods miss fewer human instances and predict more accurate alpha mattes than the instance segmentation-based matting methods.
Although SOLO + FBAMatting and QueryInst + FBAMatting have slightly higher ACC than E2E-HIM, they have much lower REC, which suggests they miss many human instances.
%


%
To qualitatively compare E2E-HIM and other methods, we visualize the estimated alpha mattes in  Figure~\ref{fig:him100}.
Specifically, we present the estimated alpha mattes of Mask R-CNN + FBAMatting, PointRend + FBAMatting, SOLO + FBAMatting, QueryInst + FBAMatting, MaskFormer + FBAMatting, and E2E-HIM on the challenge images in HIM-100K.
Note that the instance segmentation methods in the presented combinations are all trained on the HIM-100K dataset.
Human instances often overlap each other in these images, which poses a huge challenge for human instance matting.
As we can see, instance segmentation-based methods perform poorly when dealing with such images.
The predicted alpha mattes of these methods often miss limb regions such as the hands and cannot eliminate limbs of other humans well.
%
On the contrary, E2E-HIM performs much better and accurately predicts the alpha mattes of limb parts that are far away from the body and can eliminate the interference of other humans.
\begin{table}[!t]
\centering
\caption{ {Computational complexity results.}   $\dag $ denotes the computational costs that exclude the matting process. The inference speed (FPS) is calculated on an NVIDIA RTX 2080 Ti GPU with the batch size of 1. }
\begin{tabular}{l|c|c|c}
\toprule
Method & GFlops   &Params & FPS  \\
\midrule
Mask R-CNN$^\dag$~\cite{he2020mask} & 144.5  & 44.2M & 14.0  \\ %
PointRend$^\dag$~\cite{kirillov2020pointrend} &  92.6  & 60.0M& 10.4  \\ %
SOLO$^\dag$~\cite{wang2020solo} & 111.7 & 88.7M & 13.8  \\
QueryInst$^\dag$~\cite{fang2021queryinst} & 388.3  & 170.9M & 7.0 \\ %
MaskFormer$^\dag$~\cite{cheng2021maskformer} & 181.0  & 39.5M & 13.9  \\ %
\midrule
E2E-HIM &305.3& 100.8M  & 10.3\\
\bottomrule
\end{tabular}%
\vspace{-5pt}
\label{tab:eff}%
\end{table}%

To evaluate the efficiency of E2E-HIM, we summarize the computational complexity results including calculation amount, parameter amount, and inference speed (FPS) in Table~\ref{tab:eff}.
Since the computational complexity of the matting procedures of the instance-based segmentation matting methods varies with the number of human instances in the image, we only include the computational complexity results of the instance segmentation procedures.
For a human instance in a $640 \times 640$ image, FBAMatting~\cite{forte2020fbamatting} requires an additional 226 GFlops of computation, 33M of params, and 75ms of latency, and MGMatting~\cite{yu2020mask} requires additional 86 GFlops of computation, 84.8M of params, and 41ms of latency.
Compared with only the instance segmentation procedures, E2E-HIM has a similar calculation amount, parameter amount, and inference speed.
However, a complete instance segmentation-based matting method requires additional matting procedures to estimate the alpha mattes, which causes them to consume more computation and be much slower.
For example, using Mask R-CNN and FBAMatting to infer a $640 \times 640$ image with 6 instances costs 1,500 GFlops of computation and 0.5s of latency.
%
Therefore, the proposed E2E-HIM is more efficient than instance segmentation-based matting methods.

The above performance and efficiency results indicate that the proposed E2E-HIM is more efficient and effective than instance segmentation-based matting methods.
In particular, E2E-HIM has $50 \%$ lower EMSE and EMAD and is $5 \times$ faster (6 instances in a 640$\times$ 640 image) than the combination of Mask R-CNN + FBAMatting in the existing work~\cite{sun2022instmatt}.
The superiority of E2E-HIM over other methods can be attributed to two aspects:
First, E2E-HIM adopts an end-to-end framework for human instance matting to bridge the gap between the trimap generation and matting, which improves prediction accuracy.
Second, E2E-HIM introduces a united guidance network and an instance matting network to incorporate multi-instance locations and semantic correspondences into a fixed-size feature map to predict alpha mattes for multiple instances simultaneously, which improves inference efficiency.
%

\subsection{Results on Human Matting}
To evaluate the performance of E2E-HIM on the traditional human matting task, we conduct experiments on public human matting datasets. Specifically, we use PPM-100, RWP-636, and P3M to evaluate the generalization ability of E2E-HIM trained on HIM-100K. In addition, we use P3M to train and evaluate E2E-HIM on human matting.

\subsubsection{PPM-100}
We compare E2E-HIM with human matting methods such as DIM~\cite{xu2017deep}, FDMPA~\cite{zhu2017fast}, LateFusion~\cite{Zhang2019A}, SHM~\cite{chen2018semantic}, HAttMatting~\cite{2020Attention}, BSHM~\cite{liu2020boosting}, and MODNet~\cite{ke2020is}  on the PPM-100 dataset.
The quantitative results of all methods are summarized in Table~\ref{tab:ppm}.
E2E-HIM outperforms all state-of-the-art human matting methods, which suggests that E2E-HIM can be applied to the traditional human matting task.

\subsubsection{RWP-636}
We compare E2E-HIM with human matting methods such as LateFusion~\cite{Zhang2019A}, SHM~\cite{chen2018semantic}, HAttMatting~\cite{2020Attention},  MODNet~\cite{ke2020is}, GFM~\cite{li2022matting}, and P3MNet~\cite{2021Privacy} and trimap-based matting methods  such as  DIM~\cite{Zhang2019A}, GCAMatting~\cite{2020Attention}, IndexNet~\cite{chen2018semantic}, and MGMatting~\cite{yu2020mask} on the RWP-636 dataset.
The quantitative results of all methods are summarized in Table~\ref{tab:rwp}.
E2E-HIM demonstrates favorable performance against all human matting methods, underscoring its powerful generalization ability.
However, the trimap-based methods have an advantage over E2E-HIM due to the additional trimap input.
As a result, E2E-HIM only outperforms DIM and GCAMatting, while failing to outperform IndexNet and MGMatting.

\begin{table}[!t]
\centering
\caption{{Quantitative results of human matting on PPM-100~\cite{ke2020is}. }
}  
\begin{tabular}{l|c|c}
\toprule
Method & MSE   & MAD \\
\midrule
DIM~\cite{xu2017deep}   & 0.0115  & 0.0327  \\
FDMPA~\cite{zhu2017fast} & 0.0101  & 0.0178  \\
LateFusion~\cite{Zhang2019A} & 0.0094  & 0.0160  \\
SHM~\cite{chen2018semantic} & 0.0072  & 0.0158  \\
HAttMatting~\cite{2020Attention} & 0.0067  & 0.0152  \\
BSHM~\cite{liu2020boosting} & 0.0063  & 0.0137  \\
MODNet~\cite{ke2020is} & 0.0044  & 0.0087  \\
\midrule
E2E-HIM &\textbf{0.0040}  & \textbf{0.0072} \\
\bottomrule
\end{tabular}%
\label{tab:ppm}%
\end{table}%

\begin{table}[!t]
\centering
\caption{Quantitative results  of human matting on RWP-636~\cite{yu2020mask}. Trimap denotes the method adopts an auxiliary trimap input. }
\begin{tabular}{l|c|c|c|c}
\toprule
Method &Trimap& {SAD} & {MAD} & {MSE} \\
\midrule
DIM~\cite{xu2017deep} &\checkmark& 33.90 & 0.0313 & 0.0178 \\ 
GCAMatting~\cite{li2020natural}  &\checkmark& 31.09 & 0.0255 & 0.0134 \\
IndexNet~\cite{lu2019indices} &\checkmark  & 27.47 & 0.0223 & 0.0104 \\
MGMatting~\cite{yu2020mask} &\checkmark  &\textbf{27.15} & \textbf{0.0207} & \textbf{0.0066} \\
\midrule
LateFusion~\cite{Zhang2019A} &× & 71.18 & 0.0556 & 0.0423 \\
HAttMatting~\cite{2020Attention}  &× & 53.45 & 0.0376 & 0.0234 \\
SHM~\cite{chen2018semantic}  &×  & 48.95 & 0.0375 & 0.0283 \\
MODNet~\cite{ke2020is} &× & 42.63 & 0.0338 & 0.0227 \\
GFM~\cite{li2022matting}   &× & 37.93 & 0.0316 & 0.0220 \\
P3MNet~\cite{2021Privacy} &× & 36.47 & 0.0270 & 0.0183 \\
\midrule
E2E-HIM  &× & \textbf{30.33}& \textbf{0.0243} & \textbf{0.0165} \\
\bottomrule
\end{tabular}%
\label{tab:rwp}%
\end{table}%

\subsubsection{P3M}
We follow P3MNet to train E2E-HIM using the P3M dataset. Subsequently, we compare E2E-HIM with human matting methods such as LateFusion, SHM, HAttMatting, GFM, and P3MNet, on both the P3M-500-P and P3M-500-NP validation sets. 
Additionally, we evaluate the generalization capability of E2E-HIM trained on HIM-100K using the P3M-500-NP validation set. 
Note that, we sum the predictions of E2E-HIM trained on HIM-100K due to the presence of multiple instances in the images of the P3M-500-NP validation set.
The quantitative results of all methods are summarized in Tables~\ref{tab:p3ma} and~\ref{tab:p3mb}.
Our E2E-HIM outperforms all other human matting methods, demonstrating its robust scalability on human matting. 
Moreover, E2E-HIM trained on HIM-100K also outperforms most human matting methods, showcasing its good generalization ability.

\begin{table}[!t]
\centering
\caption{Quantitative results of human matting on P3M-500-P~\cite{2021Privacy}.  }
\begin{tabular}{l|c|c|c}
\toprule
Method & {SAD} & {MAD} & {MSE} \\
\midrule
LateFusion~\cite{Zhang2019A}   & 42.95&0.0250& 0.0191 \\
HAttMatting~\cite{2020Attention}  &25.99&0.0152&0.0054 \\
SHM~\cite{chen2018semantic}    &21.56&0.0125  &0.0100 \\
GFM~\cite{li2022matting} &13.20&0.0080&0.0050\\
P3MNet~\cite{2021Privacy} & 8.73&0.0051&0.0026\\
\midrule
E2E-HIM &\bf{7.96} &\bf{0.0046}&\bf{0.0024} \\
\bottomrule
\end{tabular}%
\label{tab:p3ma}%
\end{table}%

\begin{table}[!t]
\centering
\caption{Quantitative results of human matting on P3M-500-NP~\cite{2021Privacy}. * denotes the method is trained on HIM-100K. }
\begin{tabular}{l|c|c|c}
\toprule
Method & {SAD} & {MAD} & {MSE} \\
\midrule
LateFusion~\cite{Zhang2019A}   &  32.59&0.0188& 0.0131 \\
HAttMatting~\cite{2020Attention}  & 30.53 &0.0176&0.0072 \\
SHM~\cite{chen2018semantic}    & 20.77  &0.0122  &0.0093\\
GFM~\cite{li2022matting} & 15.50  & 0.0091 & 0.0056 \\
P3MNet~\cite{2021Privacy} & {11.23} & {0.0065} & {0.0035} \\
\midrule
E2E-HIM &\bf{9.25}&\bf{0.0054}&\bf{0.0030} \\
E2E-HIM* & 12.50   & 0.0072 & 0.0042\\
\bottomrule
\end{tabular}%
\label{tab:p3mb}%
\end{table}%

\begin{figure*}[!t]
\begin{center}
\includegraphics[width=0.95\linewidth]{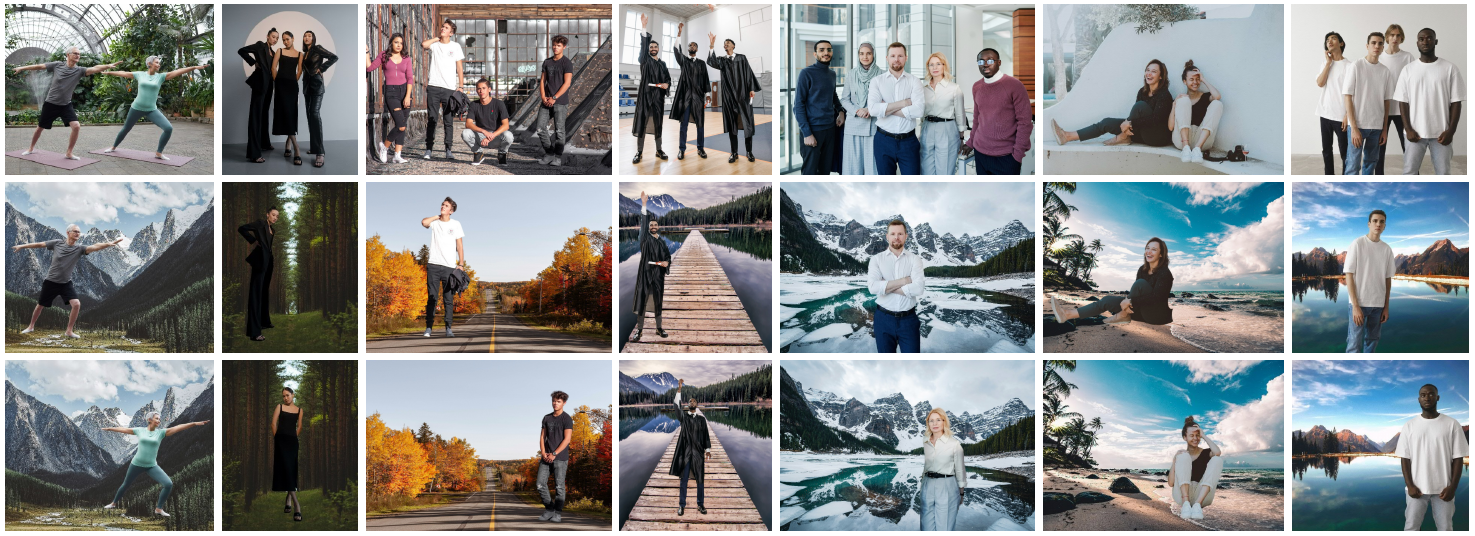}
\caption{{Examples of images synthesized with the instance alpha mattes estimated by our E2E-HIM.} The first row shows the input images. The second and third rows show the images synthesized with the estimated individual instance alpha mattes.  Zoom in for the best visualization.}
\label{fig:demos}
\end{center}
\end{figure*}

\begin{figure*}[!t]
\centering{
\includegraphics[width=0.95\linewidth]{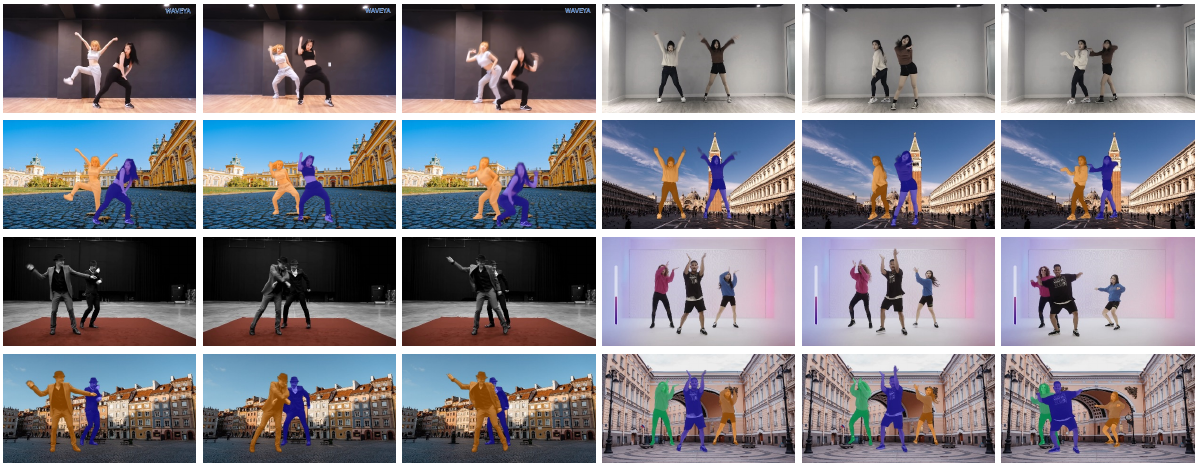}}
\caption{{Examples of video frames synthesized with the instance alpha mattes estimated by our E2E-HIM.} Each human instance in the synthesized videos are marked with a different color. Zoom in for the best visualization. }
\label{fig:real}
\end{figure*}

\subsection{Results on the Real-world Data}
\label{sec:real}
To evaluate the generalization ability of E2E-HIM on real-world data, we apply the proposed E2E-HIM to the real-world images and videos, and then subsequently generate synthetic images using  the predicted instance-level alpha mattes.

\subsubsection{Results on the Real-world Images}
To validate the generalization ability of E2E-HIM on real-world images, we apply E2E-HIM to Internet images for synthesizing new images.
Specifically, we first collect portrait images and background images from Internet galleries (e.g., pexels).
We then use E2E-HIM to estimate the alpha mattes of human instances in the collected portrait images.
Finally, we use the estimated alpha mattes to synthesize new images with the portrait images and background images.
As shown in Figure~\ref{fig:demos}, we present examples of the source images and synthesized portrait images,
which indicates that E2E-HIM can accurately estimate the alpha mattes of individual instances in the source images to synthesize natural-looking images.

\subsubsection{Results on the Real-world Videos}
To demonstrate the performance of E2E-HIM in real-world applications such as video conferencing and live streaming, we evaluate the proposed E2E-HIM on Internet videos.
Specifically, we first collect challenging dancing videos from Internet video sites (e.g., YouTube).
Then, we use E2E-HIM to estimate all instance-level alpha mattes in all frames of the videos.
Finally, we use the estimated alpha mattes to synthesize new videos with new background images.
As shown in Figure~\ref{fig:real}, E2E-HIM can accurately extract the human instances in the videos, which highlights the potential of E2E-HIM in video editing.


\subsection{Ablation study}
To evaluate the impacts of the number of guidance heads , number of queries, feature fusion, refiner, and backbones on performance,
we perform ablation studies on the HIM-100K dataset.
The quantitative results are summarized in Tables~\ref{tab:as}  and \ref{tab:addbackbone}.
Below we will analyze the results in detail.

\noindent \textbf{Guidance.}
To aggregate the context features for multi-instance matting, E2E-HIM introduces a united guidance network with a multi-head  design.
To evaluate the impacts of different guidance heads, we compare E2E-HIM,  E2E-HIM-Var1, and E2E-HIM-Var2 on the HIM-100K dataset, where E2E-HIM-Var1 has no guidance heads and E2E-HIM-Var2 has $4$ guidance heads.
We adopt the ResNet-50~\cite{he2016deep} as the backbone for all compared methods and summarize the results in Table~\ref{tab:as}.
The E2E-HIM with more heads performs better, which indicates that  the united guidance is beneficial for performance.

\begin{table*}[!t]
\centering
\caption{Ablation study on the guidance, query number, feature fusion, and refiner. All methods adopt ResNet-50~\cite{he2016deep} as the backbone.}
\begin{tabular}{l|c|c|c|c|c|c|c|c}
\toprule
Method &Head Number & Query Number & Fusion& Refiner &  $\rm{EMSE}_{0.75}$ & $\rm{EMAD}_{0.75}$ & $\rm{REC}_{0.75}$ & $\rm{ACC}_{0.75}$ \\
\midrule
E2E-HIM-Var1 & 0    & 20 & ×  & \checkmark   & 0.00663  & 0.00799  & 0.92428  & 0.90323  \\
E2E-HIM-Var2  & 4    & 20 & \checkmark& \checkmark  & 0.00592 & 0.00732 & 0.95466 & 0.92735 \\
E2E-HIM-Var3  & 2    & 10 & \checkmark& \checkmark  & 0.00642  & 0.00779 & 0.93843 & 0.91580 \\
E2E-HIM-Var4  & 2     & 50  & \checkmark& \checkmark & 0.00562 & 0.00691 & 0.95559 & 0.94065 \\
E2E-HIM-Var5    & 2 & 20 & ×   & \checkmark     & 	0.00629 & 	0.00765 &  0.93228 & 	0.91333  \\
E2E-HIM-Var6     & 2 & 20  & \checkmark & ×    & 0.00787 & 0.00944 & 0.90074 & 0.88123 \\
\midrule
E2E-HIM    & 2 & 20 & \checkmark& \checkmark   &   0.00602  & 0.00737  & 0.94863  & 0.92955  \\
\bottomrule
\end{tabular}%
\label{tab:as}%
\end{table*}%

\begin{table*}[!t]
\centering
\caption{{Ablation study on the backbone.  }   All methods adopt the query number of 20. The inference speed (FPS) is calculated on an NVIDIA RTX 2080 Ti GPU with the batch size of 1. }
\begin{tabular}{l|c|c|c|c|c|c|c|c|c}
\toprule
Backbone &  $\rm{EMSE}_{0.5}$ & $\rm{EMAD}_{0.5}$ & $\rm{REC}_{0.5}$ & $\rm{ACC}_{0.5}$ & $\rm{EMSE}_{0.75}$ & $\rm{EMAD}_{0.75}$ & $\rm{REC}_{0.75}$ & $\rm{ACC}_{0.75}$  & FPS \\
\midrule
ResNet-50 &   0.00758  & 0.00895  & 0.98562  & 0.96580  & 0.00602  & 0.00737  & 0.94863  & 0.92955 &10.3 \\
Swin-Base        & 0.00601  & 0.00731  & 0.99061  & 0.97234  & 0.00503  & 0.00632  & 0.96846 & 0.95060& 8.9 \\
Swin-Tiny     & 0.00648 & 0.00779 & 0.98864 & 0.96964 & 0.00535 & 0.00665 & 0.96185 & 0.94336&22.7 \\
ViTAEv2-S   &0.00619&0.00747&0.98875&0.98000&0.00504&0.00631&0.96451&0.95598 &9.2\\
\bottomrule
\end{tabular}%
\label{tab:addbackbone}%
\end{table*}


\noindent \textbf{Number of queries.}
To decode the instance contexts to generate latent codes, E2E-HIM follows DETR~\cite{carion2020end} to introduce the query tensor for the perception decoder.
To evaluate the impacts of the number of queries, we compare E2E-HIM, E2E-HIM-Var3, and E2E-HIM-Var4 on the HIM-100K dataset, where E2E-HIM-Var3 has 10 queries and E2E-HIM-Var4 has 50 queries.
We also adopt the ResNet-50~\cite{he2016deep} as the  backbone for all compared methods and summarize the results in Table~\ref{tab:as}.
As the number of queries increases, the performance of human instance matting improves, which suggests that a larger number of queries is beneficial to performance.
However, more queries will greatly increase the computational complexity of the general perception network and united guidance network.

\noindent \textbf{Feature fusion and refiner.}
E2E-HIM adopt the feature fusion and refiner to aggregate the context features of each instance and recover the low-level information, respectively.
To evaluate the impacts of the feature fusion and refiner, we compare E2E-HIM, E2E-HIM-Var5, and E2E-HIM-Var6 on the HIM-100K dataset, where E2E-HIM-Var5 removes the feature fusion and E2E-HIM-Var6 removes the refiner.
We also adopt the ResNet-50~\cite{he2016deep} as the  backbone for all compared methods and summarize the results in Table~\ref{tab:as}.
After removing the feature fusion and refiner, the performance of E2E-HIM on human instance matting decreases, which verifies the effectiveness of both modules.
In particular, removing the refiner hurts the prediction accuracy, which indicates that low-level image features are important to alpha matte estimation.

\noindent \textbf{Backbones.}
In Section~\ref{sec:him100}, we adopt the ResNet-50~\cite{he2016deep} as the backbone of E2E-HIM and compared methods.
To verify the scalability of E2E-HIM, we evaluate the E2E-HIM framework with the ResNet-50~\cite{he2016deep}, Swin-Tiny~\cite{liu2021Swin}, Swin-Base, and ViTAEv2-S~\cite{zhang2022vitaev2} backbones on the HIM-100K dataset.
The results are summarized in Table~\ref{tab:addbackbone}.
The proposed E2E-HIM framework achieves good performance with all four evaluated backbones.
In particular, E2E-HIM based on the Swin-Base, Swin-Tiny and ViTAEv2-S backbones outperform E2E-HIM based on the Resnet-50 backbone.
In addition, E2E-HIM using the Swin-Tiny backbone is very fast and can meet the needs of real-time human instance matting.

\section{Conclusion}
In this paper, we propose a novel End-to-End Human Instance Matting (E2E-HIM) framework for simultaneous multiple instance matting in a more efficient way.
Specifically,  a general perception network adopts a hybrid transformer to first extract image features and then decode instance contexts into latent codes.
Then, a united guidance network exploits spatial attention and semantics to generate the united semantics guidance, which encodes the locations and semantic correspondences of all human instances.
Finally, an instance matting network decodes the image features and united semantics guidance to predict all instance-level alpha mattes.
To support human instance matting research, we construct a large-scale human instance matting dataset (HIM-100K), which will facilitate future human matting research.
Extensive experiments on the HIM-100K, PPM-100, RWP-636, and P3M datasets and real-world data demonstrate the competitive performance of the proposed E2E-HIM.

\ifCLASSOPTIONcaptionsoff
  \newpage
\fi




\begin{thebibliography}{10}
\providecommand{\url}[1]{#1}
\csname url@samestyle\endcsname
\providecommand{\newblock}{\relax}
\providecommand{\bibinfo}[2]{#2}
\providecommand{\BIBentrySTDinterwordspacing}{\spaceskip=0pt\relax}
\providecommand{\BIBentryALTinterwordstretchfactor}{4}
\providecommand{\BIBentryALTinterwordspacing}{\spaceskip=\fontdimen2\font plus
\BIBentryALTinterwordstretchfactor\fontdimen3\font minus
  \fontdimen4\font\relax}
\providecommand{\BIBforeignlanguage}[2]{{%
\expandafter\ifx\csname l@#1\endcsname\relax
\typeout{** WARNING: IEEEtran.bst: No hyphenation pattern has been}%
\typeout{** loaded for the language `#1'. Using the pattern for}%
\typeout{** the default language instead.}%
\else
\language=\csname l@#1\endcsname
\fi
#2}}
\providecommand{\BIBdecl}{\relax}
\BIBdecl

\bibitem{instmattcrv}
G.~Hu and J.~J. Clark, ``Instance {S}egmentation {B}ased {S}emantic {M}atting
  for {C}ompositing {A}pplications,'' in \emph{Conference on Computer and Robot
  Vision}.\hskip 1em plus 0.5em minus 0.4em\relax IEEE, 2019, pp. 135--142.

\bibitem{he2020mask}
K.~{He}, G.~{Gkioxari}, P.~{Dollar}, and R.~{Girshick}, ``Mask {R-CNN},''
  \emph{TPAMI}, vol.~42, no.~2, pp. 386--397, 2020.

\bibitem{xu2017deep}
N.~{Xu}, B.~{Price}, S.~{Cohen}, and T.~{Huang}, ``{D}eep {I}mage {M}atting,''
  in \emph{CVPR}, 2017.

\bibitem{sun2022instmatt}
Y.~Sun, C.-K. Tang, and Y.-W. Tai, ``Human {I}nstance {M}atting via {M}utual
  {G}uidance and {M}ulti-{I}nstance {R}efinement,'' in \emph{CVPR}, 2022.

\bibitem{forte2020fbamatting}
M.~Forte and F.~Pitié, ``{F, B, Alpha Matting},'' \emph{arXiv preprint
  arXiv:2003.07711}, 2020.

\bibitem{zhu2017fast}
B.~{Zhu}, Y.~{Chen}, J.~{Wang}, S.~{Liu}, B.~{Zhang}, and M.~{Tang}, ``{F}ast
  {D}eep {M}atting for {P}ortrait {A}nimation on {M}obile {P}hone,'' in
  \emph{ACM MM}, 2017.

\bibitem{chen2018semantic}
Q.~{Chen}, T.~{Ge}, Y.~{Xu}, Z.~{Zhang}, X.~{Yang}, and K.~{Gai}, ``Semantic
  {H}uman {M}atting,'' in \emph{ACM MM}, 2018.

\bibitem{Wang_2021_ICCV}
T.~Wang, S.~Liu, Y.~Tian, K.~Li, and M.-H. Yang, ``Video {M}atting via
  {C}onsistency-{R}egularized {G}raph {N}eural {N}etworks,'' in \emph{ICCV},
  2021.

\bibitem{zhang2021attention}
Y.~Zhang, C.~Wang, M.~Cui, P.~Ren, X.~Xie, X.-S. Hua, H.~Bao, Q.~Huang, and
  W.~Xu, ``Attention-{G}uided {T}emporally {C}oherent {V}ideo {O}bject
  {M}atting,'' in \emph{ACM MM}, 2021.

\bibitem{levin2008a}
A.~{Levin}, D.~{Lischinski}, and Y.~{Weiss}, ``A {C}losed-{F}orm {S}olution to
  {N}atural {I}mage {M}atting,'' \emph{TPAMI}, vol.~30, no.~2, pp. 228--242,
  2008.

\bibitem{chen2013knn}
Q.~{Chen}, D.~{Li}, and C.-K. {Tang}, ``{KNN} {M}atting,'' \emph{TPAMI},
  vol.~35, no.~9, pp. 2175--2188, 2013.

\bibitem{lu2019indices}
H.~{Lu}, Y.~{Dai}, C.~{Shen}, and S.~{Xu}, ``Indices {M}atter: {L}earning to
  {I}ndex for {D}eep {I}mage {M}atting,'' in \emph{ICCV}, 2019.

\bibitem{li2020natural}
Y.~{Li} and H.~{Lu}, ``Natural {I}mage {M}atting via {G}uided {C}ontextual
  {A}ttention,'' in \emph{AAAI}, 2020.

\bibitem{yu2020mask}
Q.~Yu, J.~Zhang, H.~Zhang, Y.~Wang, Z.~Lin, N.~Xu, Y.~Bai, and A.~Yuille,
  ``Mask {G}uided {M}atting via {P}rogressive {R}efinement {N}etwork,'' in
  \emph{CVPR}, 2020, pp. 1154--1163.

\bibitem{Yu_2021_ICCV}
Z.~Yu, X.~Li, H.~Huang, W.~Zheng, and L.~Chen, ``Cascade {I}mage {M}atting
  {W}ith {D}eformable {G}raph {R}efinement,'' in \emph{ICCV}, 2021.

\bibitem{Zhang2019A}
Y.~Zhang, L.~Gong, L.~Fan, P.~Ren, and W.~Xu, ``A {L}ate {F}usion {CNN} for
  {D}igital {M}atting,'' in \emph{CVPR}, 2019.

\bibitem{ke2020is}
Z.~Ke, J.~Sun, K.~Li, Q.~Yan, and R.~W. Lau, ``{MODNet}: {Real-Time Trimap-Free
  Portrait Matting via Objective Decomposition},'' in \emph{AAAI}, 2022.

\bibitem{li2022matting}
J.~Li, J.~Zhang, S.~J. Maybank, and D.~Tao, ``Bridging {C}omposite and {R}eal:
  {T}owards {E}nd-to-end {D}eep {I}mage {M}atting,'' \emph{IJCV}, 2022.

\bibitem{Shen2016Deep}
X.~Shen, T.~Xin, H.~Gao, Z.~Chao, and J.~Jia, ``Deep {A}utomatic {P}ortrait
  {M}atting,'' in \emph{ECCV}, 2016.

\bibitem{2021Privacy}
J.~Li, S.~Ma, J.~Zhang, and D.~Tao, ``Privacy-{P}reserving {P}ortrait
  {M}atting,'' in \emph{ACM MM}, ser. MM '21, 2021, p. 3501–3509.

\bibitem{ren2022structure}
J.~Ren, Y.~Yao, B.~Lei, M.~Cui, and X.~Xie, ``Structure-{A}ware {F}low
  {G}eneration for {H}uman {B}ody {R}eshaping,'' in \emph{CVPR}, 2022, pp.
  7754--7763.

\bibitem{chen2022single}
B.~Chen, H.~Fu, X.~Chen, K.~Zhou, and Y.~Zheng, ``Single-image {H}uman-body
  {R}eshaping with {D}eep {N}eural {N}etworks,'' \emph{arXiv preprint
  arXiv:2203.10496}, 2022.

\bibitem{li2015deep}
J.~Li, C.~Xiong, L.~Liu, X.~Shu, and S.~Yan, ``Deep {F}ace {B}eautification,''
  in \emph{ACM MM}, 2015.

\bibitem{Velusamy_2020_CVPR_Workshops}
S.~Velusamy, R.~Parihar, R.~Kini, and A.~Rege, ``Fab{S}often: {F}ace
  {B}eautification via {D}ynamic {S}kin {S}moothing, {G}uided {F}eathering, and
  {T}exture {R}estoration,'' in \emph{CVPRW}, 2020.

\bibitem{wei2021improved}
T.~Wei, D.~Chen, W.~Zhou, J.~Liao, H.~Zhao, W.~Zhang, and N.~Yu, ``Improved
  image matting via real-time user clicks and uncertainty estimation,'' in
  \emph{CVPR}, 2021, pp. 15\,374--15\,383.

\bibitem{yang2022unified}
S.~D. Yang, B.~Wang, W.~Li, Y.~Lin, and C.~He, ``Unified interactive image
  matting,'' \emph{arXiv preprint arXiv:2205.08324}, 2022.

\bibitem{ding2022deep}
H.~Ding, H.~Zhang, C.~Liu, and X.~Jiang, ``Deep interactive image matting with
  feature propagation,'' \emph{IEEE TIP}, vol.~31, pp. 2421--2432, 2022.

\bibitem{zhang2020mask}
R.~Zhang, Z.~Tian, C.~Shen, M.~You, and Y.~Yan, ``Mask {E}ncoding for {S}ingle
  {S}hot {I}nstance {S}egmentation,'' in \emph{CVPR}, 2020.

\bibitem{wang2020solo}
X.~Wang, R.~Zhang, C.~Shen, T.~Kong, and L.~Li, ``Solo: A simple framework for
  instance segmentation,'' \emph{TPAMI}, vol.~44, no.~11, pp. 8587--8601, 2022.

\bibitem{dong2021solq}
B.~Dong, F.~Zeng, T.~Wang, X.~Zhang, and Y.~Wei, ``{SOLQ}: {S}egmenting
  {O}bjects by {L}earning {Q}ueries,'' in \emph{NeurIPS}, 2021.

\bibitem{transfiner}
L.~Ke, M.~Danelljan, X.~Li, Y.-W. Tai, C.-K. Tang, and F.~Yu, ``Mask transfiner
  for high-quality instance segmentation,'' in \emph{CVPR}, 2022.

\bibitem{he2016deep}
K.~{He}, X.~{Zhang}, S.~{Ren}, and J.~{Sun}, ``Deep {R}esidual {L}earning for
  {I}mage {R}ecognition,'' in \emph{CVPR}, 2016.

\bibitem{liu2018path}
S.~Liu, L.~Qi, H.~Qin, J.~Shi, and J.~Jia, ``Path {A}ggregation {N}etwork for
  {I}nstance {S}egmentation,'' in \emph{CVPR}, 2018.

\bibitem{bolya2019yolact}
D.~Bolya, C.~Zhou, F.~Xiao, and Y.~J. Lee, ``{YOLOACT}: {R}eal-{T}ime
  {I}nstance {S}egmentation,'' in \emph{ICCV}, 2019.

\bibitem{9367228}
X.~Zhang, H.~Li, F.~Meng, Z.~Song, and L.~Xu, ``Segmenting beyond the bounding
  box for instance segmentation,'' \emph{IEEE TCSVT}, vol.~32, no.~2, pp.
  704--714, 2022.

\bibitem{bolya2019yolact2}
D.~Bolya, C.~Zhou, F.~Xiao, and Y.~J. Lee, ``{YOLACT}++: {B}etter {R}eal-time
  {I}nstance {S}egmentation,'' \emph{TPAMI}, vol.~44, no.~2, pp. 1108--1121,
  2022.

\bibitem{chen2019tensormask}
X.~Chen, R.~Girshick, K.~He, and P.~Doll{\'a}r, ``Tensor{M}ask: {A}
  {F}oundation for {D}ense {O}bject {S}egmentation,'' in \emph{ICCV}, 2019.

\bibitem{chen2020blendmask}
H.~Chen, K.~Sun, Z.~Tian, C.~Shen, Y.~Huang, and Y.~Yan, ``Blend{M}ask:
  {T}op-down meets bottom-up for instance segmentation,'' in \emph{CVPR}, 2020.

\bibitem{10102514}
Y.~Sun, L.~Su, S.~Yuan, and H.~Meng, ``Danet: Dual-branch activation network
  for small object instance segmentation of ship images,'' \emph{IEEE TCSVT},
  pp. 1--1, 2023.

\bibitem{de2017semantic}
B.~De~Brabandere, D.~Neven, and L.~Van~Gool, ``Semantic {I}nstance
  {S}egmentation with a {D}iscriminative {L}oss,'' \emph{arXiv preprint
  arXiv:1708.02551}, 2017.

\bibitem{0SSAP}
N.~Gao, Y.~Shan, Y.~Wang, X.~Zhao, and K.~Huang, ``{SSAP}: {S}ingle-{S}hot
  {I}nstance {S}egmentation {W}ith {A}ffinity {P}yramid,'' in \emph{ICCV},
  2017.

\bibitem{fang2021queryinst}
Y.~Fang, S.~Yang, X.~Wang, Y.~Li, C.~Fang, Y.~Shan, B.~Feng, and W.~Liu,
  ``Query{I}nst: {P}arallelly {S}upervised {M}ask {Q}uery for {I}nstance
  {S}egmentation,'' \emph{arXiv preprint arXiv:2105.01928}, 2021.

\bibitem{sun2021sparse}
P.~Sun, R.~Zhang, Y.~Jiang, T.~Kong, C.~Xu, W.~Zhan, M.~Tomizuka, L.~Li,
  Z.~Yuan, C.~Wang \emph{et~al.}, ``Sparse {R}-{CNN}: {E}nd-to-{E}nd {O}bject
  {D}etection with {L}earnable {P}roposals,'' in \emph{CVPR}, 2021.

\bibitem{wang2021end}
Y.~Wang, Z.~Xu, X.~Wang, C.~Shen, B.~Cheng, H.~Shen, and H.~Xia, ``End-to-{E}nd
  {V}ideo {I}nstance {S}egmentation {W}ith {T}ransformers,'' in \emph{CVPR},
  2021.

\bibitem{cheng2021maskformer}
B.~Cheng, A.~G. Schwing, and A.~Kirillov, ``Per-{P}ixel {C}lassification is
  {N}ot {A}ll {Y}ou {N}eed for {S}emantic {S}egmentation,'' in \emph{NeurIPS},
  2021.

\bibitem{2018AlphaGAN}
S.~{Lutz}, K.~{Amplianitis}, and A.~{Smolic}, ``Alpha{G}a{N}: {G}enerative
  adversarial networks for natural image matting,'' in \emph{BMVC}, 2018.

\bibitem{tang2019learning}
J.~{Tang}, Y.~{Aksoy}, C.~{Oztireli}, M.~{Gross}, and T.~O. {Aydin},
  ``Learning-{B}ased {S}ampling for {N}atural {I}mage {M}atting,'' in
  \emph{CVPR}, 2019.

\bibitem{liu2021lfpnet}
Q.~Liu, H.~Xie, S.~Zhang, B.~Zhong, and R.~Ji, ``Long-{R}ange {F}eature
  {P}ropagating for {N}atural {I}mage {M}atting,'' in \emph{ACM MM}, 2021.

\bibitem{9634025}
Y.~Xu, B.~Liu, Y.~Quan, and H.~Ji, ``Unsupervised deep background matting using
  deep matte prior,'' \emph{IEEE TCSVT}, vol.~32, no.~7, pp. 4324--4337, 2022.

\bibitem{9197694}
F.~Zhou, Y.~Tian, and Z.~Qi, ``Attention transfer network for nature image
  matting,'' \emph{IEEE TCSVT}, vol.~31, no.~6, pp. 2192--2205, 2021.

\bibitem{10011442}
L.~Hu, Y.~Kong, J.~Li, and X.~Li, ``Effective local-global transformer for
  natural image matting,'' \emph{IEEE TCSVT}, pp. 1--1, 2023.

\bibitem{2020Attention}
Y.~Qiao, Y.~Liu, X.~Yang, D.~Zhou, and X.~Wei, ``Attention-{G}uided
  {H}ierarchical {S}tructure {A}ggregation for {I}mage {M}atting,'' in
  \emph{CVPR}, 2020.

\bibitem{10077601}
Y.~Zhou, L.~Zhou, T.~L. Lam, and Y.~Xu, ``Sampling propagation attention with
  trimap generation network for natural image matting,'' \emph{IEEE TCSVT}, pp.
  1--1, 2023.

\bibitem{peng2023rgb}
B.~Peng, M.~Zhang, J.~Lei, H.~Fu, H.~Shen, and Q.~Huang, ``Rgb-d human matting:
  A real-world benchmark dataset and a baseline method,'' \emph{IEEE
  Transactions on Circuits and Systems for Video Technology}, 2023.

\bibitem{girshickICCV15fastrcnn}
R.~Girshick, ``Fast {R-CNN},'' in \emph{ICCV}, 2015.

\bibitem{ren2015faster}
S.~Ren, K.~He, R.~Girshick, and J.~Sun, ``Faster {R-CNN}: {T}owards
  {R}eal-{T}ime {O}bject {D}etection with {R}egion {P}roposal {N}etworks,''
  \emph{NIPS}, vol.~28, 2015.

\bibitem{liu2016ssd}
W.~Liu, D.~Anguelov, D.~Erhan, C.~Szegedy, S.~Reed, C.-Y. Fu, and A.~C. Berg,
  ``{SSD}: {S}ingle {S}hot {M}ulti{B}ox {D}etector,'' in \emph{ECCV}.\hskip 1em
  plus 0.5em minus 0.4em\relax Springer, 2016, pp. 21--37.

\bibitem{redmon2016you}
J.~Redmon, S.~Divvala, R.~Girshick, and A.~Farhadi, ``You {O}nly {L}ook {O}nce:
  {U}nified, {R}eal-{T}ime {O}bject {D}etection,'' in \emph{CVPR}, 2016, pp.
  779--788.

\bibitem{carion2020end}
N.~Carion, F.~Massa, G.~Synnaeve, N.~Usunier, A.~Kirillov, and S.~Zagoruyko,
  ``End-to-{E}nd {O}bject {D}etection with {T}ransformers,'' in \emph{ECCV},
  2020.

\bibitem{vaswani2017attention}
A.~Vaswani, N.~Shazeer, N.~Parmar, J.~Uszkoreit, L.~Jones, A.~N. Gomez,
  {\L}.~Kaiser, and I.~Polosukhin, ``Attention {I}s {A}ll {Y}ou {N}eed,''
  \emph{NIPS}, vol.~30, 2017.

\bibitem{lin2020focal}
T.-Y. {Lin}, P.~{Goyal}, R.~{Girshick}, K.~{He}, and P.~{Dollar}, ``Focal
  {L}oss for {D}ense {O}bject {D}etection,'' \emph{TPAMI}, vol.~42, no.~2, pp.
  318--327, 2020.

\bibitem{milletari2016v}
F.~Milletari, N.~Navab, and S.-A. Ahmadi, ``V-net: Fully convolutional neural
  networks for volumetric medical image segmentation,'' in \emph{3DV}, 2016.

\bibitem{mhp2018}
J.~Zhao, J.~Li, Y.~Cheng, T.~Sim, S.~Yan, and J.~Feng, ``Understanding {H}umans
  in {C}rowded {S}cenes: {D}eep {N}ested {A}dversarial {L}earning and {A} {N}ew
  {B}enchmark for {M}ulti-{H}uman {P}arsing,'' in \emph{ACM MM}, 2018, p.
  792–800.

\bibitem{supervisely}
S.~Contributors, ``Supervisely person,'' \url{https://supervisely.com/}, 2021.

\bibitem{pexels}
P.~Contributors, ``Pexels.com,'' \url{https://www.pexels.com}, 2023.

\bibitem{NEURIPS2019_9015}
A.~Paszke, S.~Gross, F.~Massa, A.~Lerer, J.~Bradbury, G.~Chanan, T.~Killeen,
  Z.~Lin, N.~Gimelshein, L.~Antiga, A.~Desmaison, A.~Kopf, E.~Yang, Z.~DeVito,
  M.~Raison, A.~Tejani, S.~Chilamkurthy, B.~Steiner, L.~Fang, J.~Bai, and
  S.~Chintala, ``Py{T}orch: {A}n {I}mperative {S}tyle, {H}igh-{P}erformance
  {D}eep {L}earning {L}ibrary,'' in \emph{NeurIPS}, 2019.

\bibitem{deng2009imagenet}
J.~Deng, W.~Dong, R.~Socher, L.-J. Li, K.~Li, and L.~Fei-Fei, ``Imagenet: {A}
  large-scale hierarchical image database,'' in \emph{CVPR}, 2009.

\bibitem{he2015delving}
K.~{He}, X.~{Zhang}, S.~{Ren}, and J.~{Sun}, ``Delving {D}eep into
  {R}ectifiers: {S}urpassing {H}uman-{L}evel {P}erformance on {I}mage{N}et
  {C}lassification,'' in \emph{ICCV}, 2015.

\bibitem{loshchilov2018decoupled}
I.~Loshchilov and F.~Hutter, ``Decoupled {W}eight {D}ecay {R}egularization,''
  in \emph{ICLR}, 2019.

\bibitem{lin2014microsoft}
T.-Y. Lin, M.~Maire, S.~Belongie, J.~Hays, P.~Perona, D.~Ramanan,
  P.~Doll{\'a}r, and C.~L. Zitnick, ``Microsoft coco: {C}ommon objects in
  context,'' in \emph{ECCV}, 2014.

\bibitem{kirillov2020pointrend}
A.~Kirillov, Y.~Wu, K.~He, and R.~Girshick, ``{PointRend}: {I}mage segmentation
  as rendering,'' in \emph{CVPR}, 2020.

\bibitem{fang2021instances}
Y.~Fang, S.~Yang, X.~Wang, Y.~Li, C.~Fang, Y.~Shan, B.~Feng, and W.~Liu,
  ``Instances as {Q}ueries,'' in \emph{CVPR}, 2021.

\bibitem{liu2020boosting}
J.~Liu, Y.~Yao, W.~Hou, M.~Cui, X.~Xie, C.~Zhang, and X.-s. Hua, ``Boosting
  {S}emantic {H}uman {M}atting with {C}oarse {A}nnotations,'' in \emph{CVPR},
  2020.

\bibitem{liu2021Swin}
Z.~Liu, Y.~Lin, Y.~Cao, H.~Hu, Y.~Wei, Z.~Zhang, S.~Lin, and B.~Guo, ``Swin
  {T}ransformer: {H}ierarchical {V}ision {T}ransformer using {S}hifted
  {W}indows,'' in \emph{ICCV}, 2021.

\bibitem{zhang2022vitaev2}
Q.~Zhang, Y.~Xu, J.~Zhang, and D.~Tao, ``Vi{TAE}v2: {V}ision {T}ransformer
  {A}dvanced by {E}xploring {I}nductive {B}ias for {I}mage {R}ecognition and
  {B}eyond,'' \emph{arXiv preprint arXiv:2202.10108}, 2022.

\end{thebibliography}
%

\end{document}